\documentclass[11pt]{article} 
\usepackage{amssymb}
\usepackage[english]{babel}
\usepackage[T1]{fontenc}
\usepackage[utf8]{inputenc}
\usepackage{eurosym}
\usepackage{amsfonts}
\usepackage{amsmath}
\usepackage{graphicx}
\usepackage{float}
\usepackage{booktabs}
\usepackage{natbib}
\usepackage{ragged2e}

\usepackage[hidelinks]{hyperref}
\hypersetup{
    colorlinks=true,
    linkcolor=blue,
    citecolor=blue,
    urlcolor=blue
}
\usepackage{multirow}
\usepackage[left=2.54cm,right=2.54cm,top=2.54cm,bottom=2.54cm]{geometry} 
\usepackage[font=normalsize]{caption}
\usepackage{subcaption}
\usepackage{tabularx}
\usepackage{adjustbox}
\usepackage{array}
\newcolumntype{P}[1]{>{\centering\arraybackslash}p{#1}}
\usepackage{setspace}
\usepackage{cleveref}
\usepackage{rotating}
\usepackage[bb=dsserif]{mathalpha}
\usepackage{bm}
\usepackage{relsize}
\usepackage{amsthm}
\usepackage{enumitem}
\usepackage[ruled]{algorithm2e}
\newtheorem*{theorem-non}{Theorem}

\newtheorem*{assumption}{Assumptions}

\newtheorem{theorem2}{Theorem}

\newtheorem{definition}[theorem2]{Definition}

\def\independenT#1#2{\mathrel{\rlap{$#1#2$}\mkern2mu{#1#2}}}

\begin{document}

	\title{The Fairness of Credit Scoring Models\thanks{\baselineskip=0.9\normalbaselineskip We are grateful to Tobias Berg, Bruno Biais, Jean-Edouard Colliard, Andreas Fuster, Ansgar Walther, and to seminar participants at CREST, EM Lyon, HEC Paris, Online Corporate Finance Seminar, University of Edinburgh, University of Lorraine, and University of Nantes, as well as participants to the 14th Financial Risks International Forum, 2021 ACPR Chair Conference, 2021 Algorithmic Law and Society Symposium, 2021 French Economic Association Meeting, 2021 Hi!PARIS workshop on AI Bias and Data Privacy, 2022 AI Quality Forum, 2022 Conference on Opportunities and Risks of Digital Transformation in Finance and Beyond, 2022 Workshop on the Regulation of Algorithms in Banking and Finance, and to the 2023 Sustainable Finance Conference (Toulouse School of Economics) for their comments. We thank the ACPR Chair in Regulation and Systemic Risk, the Fintech Chair at Dauphine-PSL University, and the French National Research Agency (MLEforRisk ANR-21-CE26-0007, Ecodec ANR-11-LABX-0047, F-STAR ANR-17-CE26-0007-01) for supporting our research. 
 }
		\vspace{1cm}
		}
 	 \author{Christophe Hurlin\thanks{\baselineskip=0.8\normalbaselineskip University of Orl\'{e}ans and Institut Universitaire de France, Rue de Blois, 45067 Orl\'{e}ans, France.
 			Email: christophe.hurlin@univ-orleans.fr}  \and  Christophe P\'{e}rignon%
 		\thanks{%
 			HEC Paris, 1 Rue de la Lib\'{e}ration, 78350 Jouy-en-Josas, France. Email:
 			perignon@hec.fr} \and S\'{e}bastien Saurin\thanks{%
 			University of Orl\'{e}ans, Rue de Blois, 45067 Orl\'{e}ans, France.
 			Email: sebastien.saurin@univ-orleans.fr}\vspace{1cm}}
 	\date{\today\vspace{0.5cm}\\
 	}
    
	\maketitle
	
	\begin{abstract}
		In credit markets, screening algorithms aim to discriminate between good-type and bad-type borrowers. However, when doing so, they can also discriminate between individuals sharing a protected attribute (e.g. gender, age, racial origin) and the rest of the population. This can be unintentional and originate from the training dataset or from the model itself. We show how to formally test the \textit{algorithmic fairness} of scoring models and how to identify the variables responsible for any lack of fairness. We then use these variables to optimize the fairness-performance trade-off. Our framework provides guidance on how algorithmic fairness can be monitored by lenders, controlled by their regulators, improved for the benefit of protected groups, while still maintaining a high level of forecasting accuracy.
	\end{abstract}
	
	\vspace{0.5cm}
	\textit{Keywords: Credit markets; Discrimination; Machine Learning; Artificial intelligence \medskip }


\doublespacing
\newpage

\section{Introduction}	
	For their proponents, the growing use of Artificial Intelligence (AI) in credit markets allows processing massive quantity of data using powerful algorithms, hence improving classification between good-type and bad-type borrowers.\footnote{In this paper, we use \textit{AI} and \textit{Machine Learning} (ML) interchangeably to describe algorithms able to learn by identifying relationships within data and to produce predictive models in an autonomous manner.} This better forecasting ability leads to fewer non-performing loans and less money left on the table for algorithmic lenders. Furthermore, credit scoring algorithms permit to include small borrowers traditionally overlooked by standard screening techniques \citep{berg_fintech_2021}, can remove human biases from decision-making \citep{Howell2022}, and lead to higher responsiveness of the credit supply to demand shocks \citep{Fuster2019}.
    \medskip
	
	\noindent However, the development of AI has also stirred a passionate debate about potential discrimination biases \citep{oneil_weapons_2017,Bartlett2019}. Indeed, when automatically assessing the creditworthiness of loan applicants, credit scoring models can place groups of individuals sharing a protected attribute, such as gender, age, citizenship or racial origin, at a systematic disadvantage in terms of access (acceptance rate) or cost (interest rate). For instance, the Apple Pay app was publicly criticized for setting credit limits for female users at a much lower level than for otherwise comparable male users \citep{NYT2020}. Using detailed administrative data on US mortgages, \cite{Fuster2020} find that the use of ML algorithm increases interest rate disparity between White/Asian borrowers and Black/Hispanic borrowers. Such differences are not only detrimental for the groups being unfavorably treated, but they can also lead to severe reputation risk and legal risk for algorithmic lenders \citep{Morse2021}.\medskip
	
	\noindent How can we know whether a credit scoring model is unfair against groups of individuals that society would like to protect? If the model is shown to be unfair, what are the causes? Can we boost the fairness of this model while maintaining a high level of predictive performance? In this paper, we aim to answer these questions in three main steps. First, we utilize inference methods to test for a given null hypothesis of algorithmic fairness. Second, we propose a new interpretability technique, called \textit{Fairness Partial Dependence Plot} (FPDP), to identify the variables that generate the lack of fairness. Third, we use this set of variables to mitigate the lack of fairness problem, yet preserving the prediction accuracy of the model.\medskip
    
    \noindent Discrimination and algorithmic fairness are related yet distinct concepts. Discrimination in credit markets leads certain individuals or groups based on protected attributes having less access to credit or being offered credit on worse terms compared to others for reasons that are not related to their creditworthiness \citep{munnell1996}. Discrimination is a broad concept, which can be due to human biases, systemic inequalities or algorithmic decisions. Differently, algorithmic fairness specifically pertains to the practice of designing and implementing algorithms in a way that they operate impartially, without favoring or disadvantaging any particular group of individuals.\footnote{In this paper, we use \textit{fairness} and \textit{algorithmic fairness} interchangeably. Fairness has become a central concept for scholars working on algorithmic discrimination in ML \citep{Barocas2020}, law \citep{gillis_big_2018}, or business and economics \citep{kleinberg_algorithmic_2018,cowgill_algorithmic_2020,Kozodoi2023}.} There are many available fairness definitions, including some that neglect the creditworthiness of the applicants.\medskip

	\noindent In our analysis, we consider several definitions of fairness that appear particularly relevant in the lending context. The most commonly used definition, \textit{statistical parity}, corresponds to the equality of probability of being classified as good type in the protected and unprotected groups. Alternatively, \textit{conditional statistical parity} states that the probability of being classified as good type conditional on both displaying the protected attribute (e.g. being a woman) and belonging to a given homogeneous risk class (e.g. similar income, job tenure, marital status, and credit history) is the same in all groups. We also explore alternative definitions that take into account the true type of borrowers to account for classification errors, and definitions that rely on the predicted probability of the algorithm to eliminate dependence on a specific threshold.\medskip

	\noindent Our methodology fits nicely within the US legal framework to ensure fairness in lending, and in particular the Equal Credit Opportunity Act (ECOA) and Fair Housing Act (FHA) \citep{evans_keeping_2017, evans_catalogs_2019, bartlett_algorithmic_2020}. In credit markets, disparate impact refers to lending practices adversely affecting one group of borrowers sharing a protected attribute, even though the lender only use facially neutral variables \citep{Bartlett2019,Prince2020,Fuster2020}. Under this framework, the plaintiff must demonstrate that a lending practice impacted disparately on members of a protected group. This corresponds to our first step, namely testing the null hypothesis of equal treatment. If disparate impact has been shown, the defendant must demonstrate that the practice is consistent with business necessity. This is in line with our second step allowing us to identify the variables that lead to a fairness problem. Thus, our method can help to operationalize statistically the legal concept of fair lending.\medskip\
	
	\noindent Making sure AI algorithms treat loan applicants in a fair way is nowadays a growing concern for governments and regulators, as demonstrated by recent reports and white papers devoted to the governance of AI in finance. For instance, the EU regulation of AI approved in February 2024 states that "\textit{AI systems used to evaluate the credit score or creditworthiness of natural persons should be classified as high-risk AI systems [...] AI systems used for this purpose may lead to discrimination of persons or groups and perpetuate historical patterns of discrimination, for example based on racial or ethnic origins, disabilities, age, sexual orientation, or create new forms of discriminatory impacts}".\footnote{Other recent examples include the reports published by US \citep{BG2021} or international regulators (European Commission \citep{EC2020}, European Banking Authority \citep{EBA2020}.} However, while the issue is now universally recognized, academics, lenders, regulators, customer protection groups, lawyers and judges are still lacking tools to look at it in a systematic way. The resulting high legal and regulatory uncertainties surrounding the use of ML algorithms acts as an impediment for financial service providers to innovate and invest in screening technologies \citep{evans_keeping_2017, bartlett_algorithmic_2020}. We see our paper as an attempt to provide such tools.\medskip
	
	\noindent We illustrate our fairness assessment framework by testing for gender discrimination in a database of retail borrowers. First, we show that our tests are able to reject the fairness hypothesis when gender is explicitly used as a feature in the credit scoring model. Second, when gender is excluded from the feature space, most considered scoring models turn out to be fair regardless of the considered metrics. Interestingly, the null hypothesis of fairness is still rejected for some highly non-linear ML models. Furthermore, we observe that the choice of the parameters controlling the learning process (hyperparameters) of the algorithms strongly impacts fairness. This high sensitivity of the fairness measures to hyperparameter tuning happens to be an important source of operational risk in ML credit models. Then, we turn to the identification of the features originating the lack of fairness, called candidate variables. When scrutinizing these variables, we discover that they are of two types: those that strongly correlate with gender and/or default and those that only exhibit a weak correlation with them.\medskip
	
	\noindent We show that neutralizing a single candidate variable in the scoring model can be sufficient to achieve fairness, yet preserving the overall predictive performance of the model. In this context, neutralizing simply means using a common value for this feature for all loan applicants. It can be viewed as a post-processing method, as the scoring model is not retrained. This approach is particularly relevant in the context of credit scoring. Indeed, fairness assessment of credit scoring models typically takes place within a backtesting process that encompasses various dimensions, such as calibration, stability over time, and the homogeneity of risk grades, as conducted by internal validation teams or the supervisory authority. Within this validation assessment, the model is generally considered as given. In terms of fairness/performance trade-off, our mitigation method outperforms alternative methods based on the re-estimation of the model. As muting a single candidate variable delivers multiple fair models, we rely on a Pareto front analysis to identify optimal solutions, dominating the others \textit{both} in terms of fairness and performance (statistical or economic). Finally, we replicate our framework on a second credit scoring dataset to show that our framework can accommodate much larger datasets, on which we train the model and test for its fairness on separate samples.\medskip 
    
	\noindent This paper contributes to the literature on discrimination in lending. Our first contribution is to show how to implement inference tests for model fairness to account for estimation risk. Such tests can advantageously replace ad-hoc rules often used in practice to decide whether decisions in different groups of individuals are similar enough.\footnote{For instance, the Uniform Guidelines on Employee Selection Procedures, adopted by the Equal Employment Opportunity Commission \citep{EEOC1978} introduces the 4/5 rule as follows "\textit{a selection rate for any racial origin, sex, or ethnic group which is less than four-fifths (4/5) (or eighty percent) of the rate for the group with the highest rate will generally be regarded by the Federal enforcement agencies as evidence of adverse impact}".} The second contribution is to show how to improve the fairness of ML algorithms using a novel dedicated interpretability method to identify the variables that lead to the unfairness diagnosis. Then, we show how to treat some of these variables to remove the systematic difference between groups of applicants, without sacrificing predictive performance. 
	
	\section{Literature review}
    \label{section:lit_review}
	\noindent As this paper aims to bridge two streams of literature, we first present them separately: the financial economic literature on discrimination in Section \ref{literature_eco} and the machine-learning literature on fairness in Section \ref{literature_ml}.
 
 \subsection{Literature in financial economics} 

 \label{literature_eco}
 Discrimination in lending across demographic groups can take multiple forms: higher rejection rates, higher interest rates, lower credit limits, more guarantees, etc. Furthermore, it can happen at any stage of the life of a credit: when applying for a new loan, when refinancing it, or when asking for a credit limit extension. Several standard economic theories can explain this phenomenon. Under taste-based discrimination \citep{becker_review_1958}, some managers get utility from engaging in discrimination against individuals sharing a protected attribute, and even so when these individuals are more productive. Differently, under statistical discrimination \citep{arrow_theory_1971,phelps_statistical_1972}, firms lack information about the true creditworthiness of borrowers. To deal with this uncertainty, firms can rely on the average historical creditworthiness of each group of borrowers, or to rely on variables that are correlated with both creditworthiness and the protected attribute.\medskip
	
    \noindent There is compelling empirical evidence about discrimination in lending. Most studies analyze a dataset of actual decisions of a lender in a regression set up with information about a protected attribute (typically gender and racial origin) and features capturing the applicants creditworthiness. In this set up, a negative and statistically-significant coefficient on the protected attribute is indicative of discrimination \citep{munnell1996, calem2002anatomy, Chernenko2023}. Alternatively, if the dependent variable is the interest charge to the borrower, a positive coefficient for the protected attribute suggests discrimination \citep{alesina_women_2013, bayer_what_2018, Bartlett2019, bhutta_minorities_2021}. Other studies attribute part of the observed disparities in credit availability to a misalignment of the firm and loan officers’ incentives \citep{HERTZBERG2010, Quian2015, berg_loan_2020, dobbie_measuring_2018}.\medskip 
	
    \noindent The rise of algorithms and big data in lending has been recognized as significantly influencing the likelihood and forms of discrimination. First, employing an algorithm making objective decisions and applying similar standards to all customers can mitigate or remove discrimination based on preferences or incentives. For instance, using a sample of auto loan applicants who also requested a credit card limit increase on the same year, \cite{Butler2023} document strong racial discrimination in the auto loan market, on which decisions involve personal interaction, and no racial discrimination in the credit card market, on which decisions tend to be automated (see also \cite{Philippon2019}, \cite{dobbie_measuring_2018}, \cite{Howell2022}, and \cite{dacunto_how_2020}). Conversely, when \cite{Bartlett2019} contrast the discrimination observed for FinTech and non-FinTech lenders, they find that the rate disparities affecting minority borrowers are comparable between the two categories.\medskip
	
    \noindent Second, ML algorithms, and especially when implemented with large datasets, are likely to better capture the structural relationship between observable characteristics and default \citep{Jansen2020}. Here the effect on discrimination can go both ways. On the one hand, as shown by \cite{Berg2020}, combining advanced modeling techniques with non-standard data permits to include small borrowers traditionally overlooked by standard screening techniques. On the other hand, the model and empirical evidence in \cite{Fuster2020} indicate that ML increases rate disparity across groups of borrowers and benefits more White and Asian borrowers than Black and Hispanic borrowers. Compared to logistic regressions, ML models introduce additional flexibility which improves out-of-sample classification accuracy. However, the gains associated with this improvement are not homogeneously distributed across borrowers, as minorities may be affected by a triangulation effect. The latter occurs when non-linear associations between the features proxy for the protected variable, hence "de-anonymizing" the group identities only using non-protected attributes.\medskip
 
    \noindent While most of the studies in the literature are based on statistical measures, criteria, or causal analysis, only a few of them offer the possibility to implement formal inference procedures. For instance, the "input accountability test" proposed by  \cite{bartlett_algorithmic_2020} is a statistical test used to prevent an algorithm from systematically penalizing members of a protected group. The authors estimate a parametric model (e.g., a linear or logistic regression model) to explain the target variable with each feature, one by one. They then test the correlation between the residual and the protected attribute. Any variable with a statistically significant correlation is excluded from the decision-making model. \cite{pope2011implementing} propose a method to eliminate proxy effects while maintaining the predictive accuracy of the model. First, they estimate a regression model that includes all of the predictive features, including the protected attribute. Using this full model ensures that the predictive accuracy of the other variables does not come from their correlation with the protected attribute. Second, they only use the coefficients from the non-sensitive features to produce the individual forecasts. 

   \subsection{Literature in machine learning}
   \label{literature_ml}
   Over the years, more than twenty definitions of fairness have been put forward in the ML literature and so far, there is no consensus about which ones should be preferred (see   
   \cite{Hardt2016}, \cite{VermaRubin2018}, \cite{kleinberg_algorithmic_2018}, \cite{Lee_Floridi2020}, 
   \cite{Bono2021}, 
   or \cite{Mitchell2021}). Furthermore, some of these definitions turn out to be mathematically incompatible \citep{Kleinberg2017InherentTI,Chouldechova2017}. Among the various definitions, \cite{Barocas2020} identify three main categories or criteria: the \textit{independence} criterion refers to the independence of the protected attribute and the predicted outcome; the \textit{separation} criterion allows correlation between the score and the protected attribute to the extent that it is justified by the target variable, requiring independence within each stratum of the population defined by the target variable; and the \textit{sufficiency} criterion implies that the protected attribute and the target variable are conditionally independent given the model score.\footnote{In a recent empirical study, \cite{Kozodoi2023} compare these three definitions and recommend separation as a proper criterion for measuring the fairness of a credit scoring model, as it accounts for different misclassification costs between groups.} This categorization does not encompass fairness criteria that rely on a causal reasoning derived from a causal graph, such as \textit{counterfactual fairness}.\medskip
    
    \noindent While most of the studies in the literature are based on statistical measures, criteria, or causal analysis, only a few of them implement a formal inference procedure. For instance, \cite{kamiran2012data} implement a two-sample \textit{t} test to see whether the probability of being in the positive predicted class differs between protected and unprotected groups (independence criterion). In an adversarial framework \cite{adel2019one} modify the inputs variables and perform a two-sample maximum mean discrepancy test. Their goal is to diminish the dependence of the predictions and prediction errors on the protected attribute (independence and separation criterion).\medskip

    \noindent Concerning the origin of the lack of fairness in credit scoring, there are two predominant perspectives: one posits that it stems from the data, and the other asserts that it originates from the algorithms employed in the scoring process. A primary example of the former is the case of "selective labels" \citep{Lakkaraju2017}. Indeed, credit scoring data are labeled selectively as they depend on the choices made by human decision-makers \citep{Kleinberg_QJE2018}. For instance, we only observe the default of clients to whom a loan was granted by the credit officer and in turn, observed outcomes are not a random selection from the overall population. Recently, \cite{Coston2021} proposed a framework to address the challenges of selectively labelled data when characterizing algorithmic fairness. Another promising approach consists in using reinforcement learning methods with a specialized reward function tailored to enhance algorithmic fairness. It holds promise as a solution for mitigating biases that may arise during data collection \citep{Yang2023}.\medskip
    
    \noindent Leveraging the diverse range of statistical fairness measures and considering the origin of the lack of fairness in algorithms, several studies in the literature aim to mitigate the lack of fairness. We distinguish between three types of mitigation methods. Specifically, \textit{pre-processing methods} aim to remove bias in the training data before it is used to train the model, \textit{in-processing methods} modify the model specification and/or training process to guarantee fairness, and \textit{post-processing methods} adjust the model's predictions after it has been trained. For instance, \cite{iosifidis2019fae} propose to define group-specific thresholds to obtain predictions that satisfy a fairness criterion (see also \cite{mayson_bias_nodate}).\medskip

    \noindent Nevertheless, addressing the issue of unfairness often comes at a price, as it typically results in a decrease in the algorithm's performance \citep{Calders2009, kamiran2012decision,Calmon2017}. To investigate this trade-off, \cite{Haas2019} relies on Pareto-dominance sorting. Such an approach has been recently employed by \cite{Kozodoi2023} to compare various fairness definitions and mitigation approaches in the context of profit-oriented credit scoring using real-world data.\medskip

    \noindent To sum up, we consolidate the literature on fairness in Table \ref{Table 1 Litterature Review}.\footnote{See \cite{Caton2023} for a more comprehensive review of the fairness literature in ML.} An advantage of this table is that it permits to clearly outline our contribution. The framework we present here is \textit{fairness definition-agnostic} as it can be applied to test any type of independence, separation, and sufficiency criteria, is \textit{post-processing} as it does not require retraining the model, and is \textit{model-agnostic} as it can be applied to any credit scoring models.
    
    
\section{Measuring fairness}

\label{section:measure_fairness}
\subsection{Framework and notations} \label{Section Fairness Metrics}

We consider a bank using an algorithm to screen borrowing applications. Each applicant is associated with a binary type $Y \in\{0,1\}$. By convention, the value $1$ represents the good type, which corresponds to a favorable outcome (e.g., loan-application approval, refinancing approval, overdraft authorization). The vector $X \in \mathcal{X} \subseteq \mathbb{R}^k $ denotes the non-protected features, which include borrower features (e.g., income, assets, debt-to-income ratio, age, occupation, banking and payment data) and contract terms (e.g. loan size, loan-to-value ratio). We denote by $D \in\{0,1\}$ the \textit{sensitive} or \textit{protected} attribute (e.g. racial origin, gender, age, religion), where the value $1$ refers to an applicant belonging to the protected group and $0$ an applicant belonging to the unprotected group.\footnote{This setup can be extended to a set of $q$ sensitive attributes $D_{1},\ldots ,D_{q}$, each of them representing a specific protected group (for instance where $D_{1}$ controls for gender, $D_{2}$ for age, and so on) or representing the different values of a multinomial variable ($D_{1}$ for Asian-American borrowers, $D_{2}$ for African-American borrowers, etc.). See \cite{Mitchell2021} for a general discussion on the advantages and limits to consider combinations of attributes.}

\medskip

\noindent The goal of the bank is to build a scoring model which maps the non-protected features $X$ into a conditional probability for an applicant of being good type, $p\left( X\right) =\Pr \left( \left. Y=1\right\vert X\right) $. This probability is then transformed into a predicted outcome $\widehat{Y}$ taking a value $1$ when $p\left( X\right) $ is above a given threshold $\delta \in \left] 0,1\right[ $ and $0$ otherwise. We denote by $f\left( X\right) $ the classifier mapping $X$ into $\widehat{Y}$, with $\widehat{Y}=f\left( X\right) $. We impose no constraint on the function $f$: it can be parametric (logistic regression for instance) or not, linear or not, an individual or ensemble classifier which combines a set of homogeneous or heterogeneous models, etc. Finally, to comply with current banking regulation, we assume that the bank does not use the sensitive attribute $D$ as an input in its scoring model.

\subsection{Fairness definitions}
	
\noindent As shown in Section \ref{literature_ml}, the literature on designing fair algorithms is extensive and interdisciplinary. Here, we focus on four fairness definitions that are representative of the main categories identified by \cite{Barocas2020}, namely independence, separation, and sufficiency, and which are particularly relevant for credit scoring models.

\begin{definition}
    A credit scoring model satisfies the independence property (statistical parity) if the predicted outcome and the sensitive attribute are independent, i.e., if $\widehat{Y}\mathpalette{\protect\independenT}{\perp}D$.
\end{definition}

\noindent The core concept of statistical parity is that all applicants should have an equivalent opportunity to obtain a good outcome from the credit scoring model, regardless of their group membership. Expressed in terms of conditional probabilities, statistical parity implies that $\Pr(\widehat{Y}=1\text{\/}|\text{\/}D=1)=\Pr (\widehat{Y}=1\text{\/}|\text{\/}D=0)=\Pr (\widehat{Y}=1)$. While it is straightforward and intuitive, this criterion comes with certain limitations. In practice, differences in outcomes between protected and unprotected groups may not be due to discrimination but, for instance, to compositional differences in the groups themselves. In turn, it is more informative to compare similar applicants from protected and unprotected groups, focusing on their comparable individual features (income, employment, education, etc.). Such an approach aligns with the concept of conditional statistical parity.
	
\begin{definition}
    A credit scoring model satisfies the conditional statistical parity property if the predicted outcome and the sensitive attribute are independent, controlling for a subset of non-protected attributes $X_{c}\subseteq X,$ i.e., if $\widehat{Y}\mathpalette{\protect\independenT}{\perp}D|X_{c}$.
\end{definition}
	
\noindent The logic consists in comparing the model outcomes for protected and unprotected group members who share the same characteristics $X_c$. This definition raises the issue of the choice of the conditioning attributes. The goal is to control for the variables that are known to have a first-order impact on creditworthiness in order to constitute homogeneous risk classes. Alternatively, one can rely on a clustering algorithm to partition the individuals into the risk classes, which has the advantage of not selecting the important variables ex-ante. The statistical parity diagnosis can be carried out in each class separately or for all classes, using an aggregation rule.

\medskip

\noindent While statistical parity is based on the joint distribution of $(\widehat{Y},D)$, other fairness definitions also take into account the true type $Y$ of the borrowers. By considering the difference between realized and predicted outcomes for protected and unprotected groups, these approaches permit to test whether there are some difference of treatment across groups in terms of classification errors.
    
\begin{definition}
    A model satisfies the separation property (equal odds), if the predicted outcome $\widehat{Y}$ and the protected attribute $D$ are independent conditional on the actual outcome $Y,$ i.e., if $\widehat{Y}\mathpalette{\protect\independenT}{\perp}D|Y$.
\end{definition}
	
\noindent Unlike statistical parity, equal odds allows $\widehat{Y}$ to depend on $D$ but only through the target variable $Y$. It implies that applicants with a good credit type and applicants with a bad credit type should have similar classification, regardless of their (protected or unprotected) group membership. Thus, a credit scoring model is considered fair if the predictor has equal True Positive Rates (TPR, i.e., probability of the truly positive subject to be identified as such) and equal False Positive Rates (FPR, i.e., probability of falsely accepting a negative case). This implies the following constraint:
\begin{equation}
    \Pr (\widehat{Y}=1\text{\/}|\text{\/}Y=y,D=0)=\Pr (\widehat{Y}=1\text{\/}|%
	\text{\/}Y=y,D=1)=\Pr (\widehat{Y}=1\text{\/}|\text{\/}Y=y),~~~y\in \left\{
		0,1\right\}
\end{equation}%
\noindent A possible relaxation of equalized odds is to require non-discrimination only within the positive-outcome group. That is, to require that people who are actually good type, have an equal opportunity of getting the loan in the first place. This relaxation is often called \textit{equal opportunity.} Formally, it implies $\left. \widehat{Y}\mathpalette{\protect	\independenT}{\perp}D\right.\text{\/}|\text{\/}Y=1$ and the equality of the TPRs for protected and unprotected groups, meaning that $\Pr (\widehat{Y}=1\text{\/}|\text{\/}Y=1,D=0)=\Pr (\widehat{Y}=1\text{\/}|\text{\/}Y=1,D=1)$. Conversely, a second relaxation called \textit{predictive equality}, reflects the equality of the FPRs for both groups, meaning that $\Pr (\widehat{Y}=1\text{\/}|\text{\/}Y=0,D=0)=\Pr (\widehat{Y}=1\text{\/}|\text{\/}Y=0,D=1)$,
and corresponds to the assumption $\widehat{Y}\mathpalette{\protect\independenT}{\perp}D|Y=0$. Unlike (conditional) statistical parity, a model satisfying equal odds may exhibit different approval rates for the protected and unprotected groups. This is particularly suitable when the protected attribute is actually correlated with the target variable.
\medskip

\noindent Furthermore, fairness definitions can be extended to the score $p(X)$ itself. 

\begin{definition}
    A credit scoring model satisfies the sufficiency property, if the actual outcome $Y$ and the protected attribute $D$ are independent conditional on the score $p(X), $ i.e., if $Y\mathpalette{\protect\independenT}{\perp}D|p(X)$.
\end{definition}

\noindent The main idea is that applicants receiving the same model-predicted score $p(X)$ should have the same probability of being good type ($Y=1$), regardless of their group membership. Formally, a score is said sufficient at a threshold $\delta$, if $\Pr(Y=1|D=1,p(X)>\delta) = \Pr(Y=1|D=0,p(X)>\delta)$. In practice, banks often utilize homogeneous risk classes derived from the credit scores. Within this framework, an algorithm fulfills the sufficiency property if all individuals within a risk class $C=c_k$ have an equal probability of being good type ($Y=1$), regardless of their group membership, implying:
\begin{equation}
    \Pr(Y=1|D=1,C_k) = \Pr(Y=1|D=0,C_k) = \Pr(Y=1|C=c_k)
\end{equation}
By design, most ML models aim to achieve sufficiency, which will be satisfied if the protected attribute can be accurately estimated using other variables \citep{Liu2019}. This is why, sufficiency is not a strict fairness condition. 
\medskip

\noindent At this stage, it is important to make a clear distinction between (1) the existence of significant differences in approval rates or interest rates, between protected and unprotected groups and (2) the notions of lack of fairness or discrimination. First, all fairness definitions, except statistical parity, are compatible with such significant differences. For instance, a model can be fair according to the equal odds definition, yet delivering different approval rates for men and women. Second, imposing fairness according to statistical parity (i.e., imposing the same approval rates) can generate discrimination for some members of the unprotected group. Indeed, in this case, two individuals with the same creditworthiness, but who are not members of the same group, may end up with different outcomes. It could also lead to more defaults in the protected group, and the total cost (economic, psychological, etc.) of this default is likely to dwarf the opportunity cost associated with their application being turned down \citep{Kozodoi2023}.
\medskip

\noindent Another issue arises when the protected attribute is correlated with default. In this case, imposing conditional statistical parity may disadvantage the \textit{protected} group. A well-known example is auto insurance pricing: conditional statistical parity implies that the pricing must be equal for men and women, as long as they display the same driving history and other observable features. However, it does not take into account the fact that gender may be correlated with the target variable, say auto insurance claims, even after accounting for observables. It is indeed well established that women are less likely to experience severe accidents, all other things being equal. As a result, enforcing conditional statistical parity based on gender can be unfair to women, causing them to pay more than they should. In this case, equal odds or equal opportunity should be preferred, as they allow for disparities in acceptance rates or pricing. 

\section{Fairness diagnosis}
	
\subsection{Fairness inference}
\label{Section_Fairness_Inference}
	
\noindent When the joint distribution of the random variables $(\widehat{Y},Y,D)$ is known, one can determine without ambiguity whether a credit scoring model satisfies any fairness definition. However, in practice, one needs to rely on empirical distributions. Here, we propose a general testing methodology that considers estimation uncertainty to statistically test for the fairness of a credit scoring model. We consider a test sample $S_{n}$ of $n$ observations $\{y_{j},x_{j},d_{j}\}_{j=1}^{n}$ issued from the joint distribution $p_{Y,X,D}$ where index $j$ denotes the $j^{th}$ credit applicant. For this sample, the credit scoring model produces a set of decisions $\{\widehat{y}_{j}\}_{j=1}^{n}$. Considering the sample $\{\widehat{y}_{j},y_{j},d_{j}\}_{j=1}^{n}$, we wish to test whether the model satisfies a particular fairness definition indexed by $i\in \left\{SP,CSP,EO,EOP,PE,SUF\right\}$ with:\vspace{0.7cm}

\begin{tabular}{p{5.1cm}p{5.1cm}p{5.1cm}}
$\text{H}_{0,SP}:\widehat{Y}\mathpalette{\protect\independenT}{\perp}D$ & $\text{H}_{0,CSP}:\widehat{Y}\mathpalette{\protect\independenT}{\perp} D|X_{c}$ & $\text{H}_{0,EO}:\widehat{Y}\mathpalette{\protect\independenT}{\perp}D|Y$ \vspace{0.2cm} \\ 
$\text{H}_{0,EOP}:\widehat{Y}\mathpalette{\protect\independenT}{\perp} D|Y=1$ & $\text{H}_{0,PE}:\widehat{Y}\mathpalette{\protect\independenT}{\perp}D|Y=0$ & $\text{H}_{0,SUF}:Y\mathpalette{\protect\independenT}{\perp}D|p(X)$ \\
\end{tabular} \vspace{0.25cm}

\noindent where $SP$ stands for statistical parity, $CSP$ for conditional statistical parity, $EO$ for equal odds, $EOP$ for equal opportunity, $PE$ for predictive equality, and $SUF$ for sufficiency. Formally, we denote a fairness test statistic as: 
	\begin{equation}
		F_{H_{0,i}}\equiv h_{i}(\widehat{Y}_{j},Y_{j},D_{j};j=1,\ldots
		,n)=h_{i}\left( f\left( X_{j}\right) ,Y_{j},D_{j};j=1,\ldots ,n\right)
	\end{equation}%
\noindent where $h_{i}\left( .\right) $ denotes a functional form that depends on {\normalsize the null hypothesis }H$_{0,i}${\normalsize, the scoring model }$f(X)$, and the sample $\{\widehat{y}_{j},y_{j},d_{j}\}_{j=1}^{n}$. As all fairness metrics can be expressed in terms of (conditional) independence assumptions, we can derive specific inference. Under the null hypothesis, the test statistic $F_{H_{0,i}}$ has a $\mathcal{F}_{i}$ distribution, and we denote $d_{1-\alpha }$ the corresponding critical value at the $\alpha \%$ significance level. An inference procedure offers several advantages. First, regardless of the chosen test statistic, fairness inference enables us to account for estimation uncertainty when comparing conditional probabilities for protected and unprotected groups. Second, it allows us to set the probability of incorrectly concluding that a scoring model is unfair, through the significance level of the test. Third, fairness test statistics can aggregate the fairness diagnoses obtained in all classes of credit applicants, without making ad-hoc assumptions about the number of classes in which the fairness assumption can be rejected or not.

\medskip
	
\noindent The notation for $F_{i}$ encompasses a wide class of test statistics that can be implemented in this context. For instance, one can use the chi-squared conditional independence test, the Cochran-Mantel-Haenszel (CMH) test, the z-tests associated with the hypothesis tests of the difference, ratio, or odds ratio of two independent proportions (for the size and power properties of these tests in finite sample, see \cite{Shah2020} or \cite{Fay2021}). In the sequel, we formally present a likelihood-ratio (LR) test that can be applied to any null hypothesis of fairness.

\medskip

\noindent Denote by $A$, $B$, and $C$, the variables of interest for a given fairness definition, where $(A,B) \in \left\{0,1\right\}^2$ are the two binary variables for which we test independence, and $C\in\{c_{1},\dots,c_{k},\dots,c_{K}\}$ is the conditioning variable. We wish to test whether the credit scoring model satisfies a particular fairness definition expressed as 
a null hypothesis $\text{H}_{0,i}: A\mathpalette{\protect\independenT}{\perp}B|C$ with $i\in \left\{SP,CSP,EO,EOP,PE,SUF\right\}$. 
For instance, testing the null hypothesis H$_{0,EO}$ of equal odds implies $A=\widehat{Y},$ $B=D,$ and $C=Y$, with $K=2$, and $\{c_1,c_2\}=\{0,1\}$. This notation encompasses most of the fairness metrics used in the literature such as statistical parity ($A=\widehat{Y}$, $B=D$, and $C=\varnothing$), conditional statistical parity ($A=\widehat{Y}$, $B=D$, and $C=X_{c}$), predictive equality ($A=\widehat{Y}$, $B=D$, and $C=\left\{ Y=0\right\}$), equal opportunity ($A=\widehat{Y}$, $B=D$, and $C=\left\{ Y=1\right\}$), and sufficiency with $A=Y,$ $B=D,$ and $C=c_k$ for $k=1,\dots,K$, where $c_k$ refers to a risk class constructed from the continuous score $p(X)$.

\medskip

\noindent Without loss of generality, we rewrite the null hypothesis of fairness as $H_{0,i}:p_{u,v,k}=\alpha_{u,k}\beta_{v,k}$ for $k= 1,...,K$, and $(u,v)\in\{0,1\}^2$, with $p_{u,v,k}= \Pr ((A=u) \cap (B=v)|C=c_k)$, $\alpha_{u,k}= \Pr (A=u|C=c_k)$, and $\beta_{v,k}= \Pr (B=v|C=c_k)$. Denote by $p_{u,v,k}(\theta_k)=\alpha_{u,k}\beta_{v,k}$, the probability under the null hypothesis for the class $c_k$, with $\theta_k = (\alpha_{1,k}, \beta_{1,k})$ and $\theta=(\theta_1, \dots, \theta_K) \in \Theta$ the vector of free parameters, with $q=\max(K,1)$, and $\Theta=[0,1]^{2q}$. For each class, the bank considers a $2 \times 2$ contingency table for $(A,B)$ with the frequencies $n_{u,v,k}=\sum_{i=1}^{n} \mathbb{1} ((A_i=u) \cap (B_i=v)|C=c_k)$, $\sum_{u=0}^{1}\sum_{v=0}^{1} n_{u,v,k}=n_k $ and $\sum_{k=1}^{K}n_k=n$, where $\mathbb{1}(.)$ denotes the indicator function, and $n_k$ corresponds to the number of individuals in the class $c_k$. Below, we make four mild assumptions concerning the true value of $\theta_k$ and $p(\theta_k)=(p_{0,0,k}(\theta_k),p_{0,1,k}(\theta_k),p_{1,0,k}(\theta_k),p_{0,0,k}(\theta_k))$ so that Taylor expansions can be made in the neighborhoods of $\theta$.
\begin{assumption}
    \label{Assumptions1} The parameters $\theta_k \in \Theta$ and $p(\theta_k)$ satisfy the following conditions:
    \begin{enumerate}[topsep=2pt, partopsep=2pt]
        \item $\theta_k$ is not on the boundary of $\Theta$.
        \item All $p_i > 0$. 
        \item $\partial p_i(\theta_k) / \partial \theta_{k,j}$ is continuous in a neighborhood of $\theta_k$.
        \item The matrix $A = \{ \partial p_i(\theta_k) / \partial \theta_{k,j} \}$ has full rank $q$ at $\theta_k$. 
    \end{enumerate}
\end{assumption}

\noindent Under these assumptions, the fairness test can be defined as a general Likelihood-Ratio test of $H_{0,i}:p_{u,v,k}=p(\theta_k)$ for $k= 1,...,K$. 
  
\begin{theorem-non}[Fairness test]
    \label{Distribution}
     Under the null hypothesis of fairness $\text{H}_{0,i}$, the test statistic $F_{H_{0,i}}$ converges in distribution to a chi-squared distribution as the sample size $n$ tends to infinity:
    \begin{align*}
        F_{H_{0,i}}=\sum_{k=1}^{K} \sum_{u=0}^{1} \sum_{v=0}^{1}
        \frac{(n_{u,v,k}-n_kp_{u,v,k}(\hat{\theta}))^2}{n_kp_{u,v,k}(\hat{\theta})}  
        \underset{H_{0,i}}{\overset{d}{\longrightarrow}} \chi^2(q)
    \end{align*}
\end{theorem-non}
\noindent The proof, which is reported in Appendix A, is derived from \cite{Seber2013}. The test represents a generalization of the Pearson's chi-squared test of independence, enabling a relatively straightforward implementation. When evaluating statistical parity ($C=\varnothing$ and $K=0$), the test reduces to the conventional chi-squared test of independence. In the general case, the null hypothesis $H_{0,i}$ is rejected at a significance level $\alpha \in \left] 0,1\right[$ whenever the test statistic $F_{H_{0,i}}$ exceeds the $1-\alpha$ quantile of the $\chi ^{2}\left(q\right)$ distribution, denoted by $d_{1-\alpha}$. Additionally, it is possible to conduct a comprehensive analysis by evaluating the null fairness assumption for a specific class $c_k$.

\subsection{Interpretability}
	
\label{Sous_Section_FPDP}

\noindent Interpretability is at the heart of the financial regulators' current concerns about the governance of AI, especially in the credit scoring industry (Consumer Financial Protection Bureau \citep{CFPB_2022} and \cite{EBA2021}). Here, we define interpretability as the degree to which a human can understand the determinants of a decision \citep{Miller2019}. It is important to note that interpretability does not necessarily implies fairness. Conversely, non-interpretable models can be fair.
\medskip

\noindent There are two large family of ML models: the natively interpretable (white box) models vs. the uninterpretable (black box) models. To allow interpreting predictions of black box models, various model-agnostic methods have been developed (see \cite{molnar2019} for an overview). For instance, the \textit{Partial Dependence Plot} (PDP) displays the marginal impact of a specific feature on the outcome $\widehat{Y}$ of a model. This plot is useful to explore whether the relationship between the target and a feature is linear or more complex. The PDP method has the advantage of being model-agnostic in the sense that it permits to explain the predictions of any ML model independently of its form, options, or internal model parameters.
\medskip
	
\noindent Instead of focusing on the interpretability of the outcome of the model, we focus on the interpretability of its fairness diagnosis. We define fairness interpretability as the degree to which an applicant, a regulator, or any stakeholder, can understand the determinants of the lack of fairness with respect to a given protected attribute induced by a ML model. Here, we propose a simple model-agnostic method, which we call \textit{Fairness Partial Dependence Plot} (FPDP). The latter displays the marginal effect of a specific feature on a fairness diagnosis test associated with a credit scoring model. The logic of the FPDP is similar to that of PDP, except that we explore the relationship between one feature and a fairness test statistic. The goal of the FPDP is to identify the feature(s) at the origin of the lack of fairness, whatever the hypothesis considered. These features are called \textit{candidate variables}.
\medskip
	
\noindent We denote by $X_{A}$ the feature for which we want to measure the marginal effect, $X_{B}$ the other features, and $f\left( .\right) $ the credit scoring model such that $\widehat{Y}=f\left(X_{A},X_{B}\right) $. Then, we can rewrite the fairness test statistic $F_{H_{0,i}}$ as:
\begin{equation}
		F_{H_{0,i}} \equiv h_{i}(\widehat{Y}_{j},Y_{j},D_{j};j=1,\ldots ,n) =\widetilde{h}_{i}(X_{A,j},X_{B,j},Y_{j},D_{j};j=1,\ldots ,n)
\end{equation}
\noindent with $\widetilde{h}\left( .\right) $ a nonlinear positive function. We have to distinguish two cases depending on whether $X_{A}$ is a categorical feature or a continuous feature.
	
\begin{definition}
    Consider a categorical feature $X_{A}\in $ $\left\{a_{1},\ldots,a_{p}\right\} $, the corresponding FPDP is:
	\begin{equation}
			F_{H_{0,i}}\left( a_{s}\right) =\widetilde{h}_{i}\left(
			(a_{s},x_{B,1},y_{1},d_{1}),\ldots ,(a_{s},x_{B,n},y_{n},d_{n})\right)
			~~~\forall s=1,\ldots ,p 
	\end{equation}
	Consider a continuous feature $X_{A}\in $ $\left[ x_{A}^{\min },x_{A}^{\max }\right] $, the corresponding FPDP is defined as:
	\begin{equation}
	F_{H_{0,i}}\left( x\right) =\widetilde{h}_{i}\left(
			(x,x_{B,1},y_{1},d_{1}),\ldots ,(x,x_{B,n},y_{n},d_{n})\right) ~~~\forall
			x\in \left[ x_{A}^{\min },x_{A}^{\max }\right]
	\end{equation}
 \end{definition}
	
\noindent The FPDP displays the realization of the fairness test statistic obtained when the categorical feature $X_{A}$ takes a value $a_{1}$, $a_{2}$, \dots, $a_{p}$ for all instances, whatever their other features. In the continuous case, the FPDP associated with the value $x$ corresponds to the fairness test statistic obtained when $X_{A}$ takes the value $x$ for all instances, whatever their other features. For instance, if the age of the applicants goes from 25 to 62, we compute the test statistics by setting the age of all applicants to 25 years, then to 26 years, \dots, until 62 years. Obviously, the FPDP depends on the null hypothesis H$_{0,i}$ which is considered. Thus, we can for instance define a FPDP with respect to (conditional) statistical parity, equal odds, equal opportunity, predictive equality, or sufficiency. \medskip
	
\noindent Intuitively, FPDP allows us to identify the candidate variables that contribute to the rejection of the fairness null hypothesis. Regardless of the feature's type (continuous or categorical), consider a case for which the null of fairness is initially rejected as $F_{H_{0,i}}>d_{1-\alpha}$ where $d_{1-\alpha}$ is the critical value of the test for $\alpha \%$ significance level associated with the null distribution $\mathcal{F}_{i}$, typically a chi-squared distribution for the LR test presented earlier. A feature is identified as a candidate variable if changing its value reverses the fairness diagnosis. 
	
\begin{definition}
	A feature $X_{A}$ is considered as a candidate variable, if there exists at least one value $a_{s}\in \left\{ a_{1},\ldots ,a_{p}\right\} $ (categorical variable) or $x\in \left[ x_{A}^{\min },x_{A}^{\max }\right] $ (continuous variable), such that $F_{H_{0,i}}\left( a_{s}\right) <d_{1-\alpha}$ or $F_{H_{0,i}}\left( x\right) <d_{1-\alpha}$.
\end{definition}

\noindent The trick of FPDP lies in identifying candidate variables by severing the connection between the features and the protected attribute. This is crucial because fairness violations often stem from features collectively acting as surrogates for the protected attribute. For example, combining information about an applicant's income, sector, and employment history can potentially reveal gender. In such cases, setting the value of one of these features can impede the model from generating a proxy for the protected attribute, effectively breaking the link. This feature is then identified as a candidate variable.  
\medskip	

\subsection{Mitigation}
\label{section:Mitigation}
    
\noindent The FPDP serves as an effective instrument for addressing the problem of unfairness. Indeed, a natural approach consists in neutralizing the effect of a candidate variable by setting its value to a reference point for which the null hypothesis of fairness cannot be rejected anymore, i.e., $F_{H_{0,i}}<d_{1-\alpha}$. Interestingly, setting the value of one candidate variable is akin to slightly modify the model. For instance, in a logistic regression model, it can be interpreted as a change in the constant term or, similarly, as a change in the probability threshold splitting applicants between good and bad types. For a classification tree, it entails removing certain subtrees that involve the candidate variable. In practice, multiple candidate variables can be identified using FPDP, and for each candidate variable, multiple values can be used to reach an equity diagnosis. This leads to a set of fairs models, i.e., a combination of candidate variable and a particular value. This is a key advantage of our approach, as it gives us many degrees of freedom to mitigate the fairness issue.  

\medskip

\noindent One strategy is to select among the fair models, the one with the highest accuracy, as measured for instance by AUC. This aligns with the general discussion of the trade-off between fairness and predictive performance \citep{Haas2019}. Several studies on algorithmic fairness have demonstrated that enhancing fairness can deteriorate overall accuracy. As our mitigation method picks among a set of fair models the one with the highest accuracy measure, it minimizes by design the drop in accuracy. Below, we provide a pseudo algorithm that describes our method, where $C_A=\{X_A^{(1)}, \hdots, X_A^{(K)}\}$ denotes the set of candidate variables identified by the FPDP.

\begin{algorithm*}[h!]
    \caption{Mitigation method and optimal model}\label{alg:neutralization}
    \KwData{$\mbox{features } X, \mbox{estimated model }\hat{Y}=f(X)$, $\mbox{set of candidate variables }C_A,$ $\mbox{sample }(x_j,y_j,d_j)_{j=1}^n$, $\mbox{fairness definition }i \in \{SP,CSP,EO,PE,EOP,SUF\}$, $\mbox{critical value } d_{1-\alpha}$, $\mbox{accuracy measure } accuracy(\hat{y},y)$}
    \KwResult{$\mbox{Optimal model }M$, $\mbox{Optimal accuracy }P$}
    $P \gets 0$
    
    \For{$k$ in $card(C_{A})$}{
        Define $X_B^{(k)}=X\setminus X_A^{(k)}$ 
        
        \For{$x$ in $[min(X^{(k)}_A),max(X^{(k)}_A)]$}{
        s $\gets 0$
    
        Define $x^{(k,s)}_A=x$
        
        Compute  $\hat{y}^{(k,s)}_j=f(x^{(k,s)}_A,x_{B,j}^{(k)}) ~ \forall j=1,\hdots,n$
        
        Compute $F_{H_{0,i}}^{(k,s)}=h_i(\hat{y}^{(k,s)}_j,y_j,d_j;j=1,\hdots,n)$
    
        \If{$F_{H_{0,i}}^{(k,s)}<d_{1-\alpha}$}{
    
            Compute $accuracy(\hat{y}^{(k,s)},y)$

            \If{$accuracy(\hat{y}^{(k,s)},y) > P$}{
            
            Define $\hat{Y}^{(k,s)} =f(x^{(k,s)}_A,X_B^{(k)})$
            
            $M \gets \hat{Y}^{(k,s)}$

            $P \gets accuracy(\hat{y}^{(k,s)},y)$
            
            s $\gets s + 1$
        } }      
    }}
\end{algorithm*}

\medskip

\noindent Alternatively, instead of relying on a fair/unfair binary diagnosis, one could rely on the \textit{degree of fairness}, as measured by the fairness test statistics $F_{H_{0,i}}$. This optimization can be represented using a Pareto analysis in the fairness-accuracy space. Pareto-dominance sorting relies on the concepts of dominated and non-dominated solutions to identify favorable trade-offs. For instance, solution A is considered dominated by solution B if B is better than or equal to A in all objective values (fairness and accuracy) and strictly better in at least one objective. The collection of Pareto-efficient solutions forms the boundary, also termed the Pareto front, which represents the optimal combinations of the multiple objectives that the algorithm can achieve. 

\medskip

\noindent Any mitigation method inevitably alters the accuracy of the credit scoring model, but also the bank's costs, profits, and number of loans granted. This discussion suggests another potential trade-off that could arise between fairness and economic performance. Our mitigation method can be adapted to find an optimal model based on this trade-off. To do so, we need to replace the accuracy metric by an economic criteria, such as the bank cost or profit. 

\medskip    

\noindent A natural alternative to our post-processing mitigation method could be to remove each candidate variable one at a time and re-estimate the scoring model. However, there is no guarantee that re-estimating the model with one less variable will successfully address fairness issues. Indeed, the retrained model might still contain features closely linked to the variable that was removed. Additionally, in real-world situations, fairness assessments of credit scoring models are usually done during backtesting. The latter encompasses a series of tests, such as calibration, stability over time, and the homogeneity of risk grades. These tests are typically carried out by internal validation groups or banking regulators. In these evaluations, the model is generally considered as given, making any re-estimation of the model infeasible in practice. 

\section{Application}
	
	\subsection{Data}
	
	\noindent We illustrate our methodology using the German Credit Dataset, which includes 1,000 consumer loans extended to respectively 310 women and 690 men.\footnote{See \cite{VermaRubin2018} and \cite{UCI} for details. In the initial database, the gender and the marital status of the applicants are specified in a common attribute with five categorical values (single male, married male, divorced male, single female, married or divorced female). Here, as we focus on gender discrimination whatever the marital status, we consider a binary variable representing the single, married, or divorced females (protected group) versus the single, married, or divorced males (unprotected group). The original database can be found \href{https://archive.ics.uci.edu/ml/datasets/statlog+(german+credit+data)}{here}.} For each applicant, we know his or her actual credit-risk type, called \textit{credit risk}: good type or low risk ($Y=1$) vs. bad type or high risk ($Y=0$). This variable is our target variable. In total, 300 borrowers are in default ($Y=0$) among which 191 are men and 109 women. Moreover, there are 19 explanatory variables measuring either some attributes of the borrower (e.g., gender, age, occupation, credit history) or some characteristics of the loan contract (e.g., amount, duration). More information about the database can be found in Tables \ref{TabX} and \ref{feature_overview} in the Appendix. We contrast in Figure \ref{Figure1 Features Distributions} in the Appendix the distributions of each of the 20 variables for men and women. We see that the default rate is higher for women ($35.16\%$) than for men ($27.68\%$). Moreover, women borrowers tend to be younger and to exhibit a lower home-ownership rate and employment duration.

	\medskip
	
\begin{figure}[!htbp] 
		\centering 
		\caption{Measures of association between features, target variables, and gender} 
		\includegraphics[width=1\textwidth,trim={0cm 1cm 0 2cm},clip]{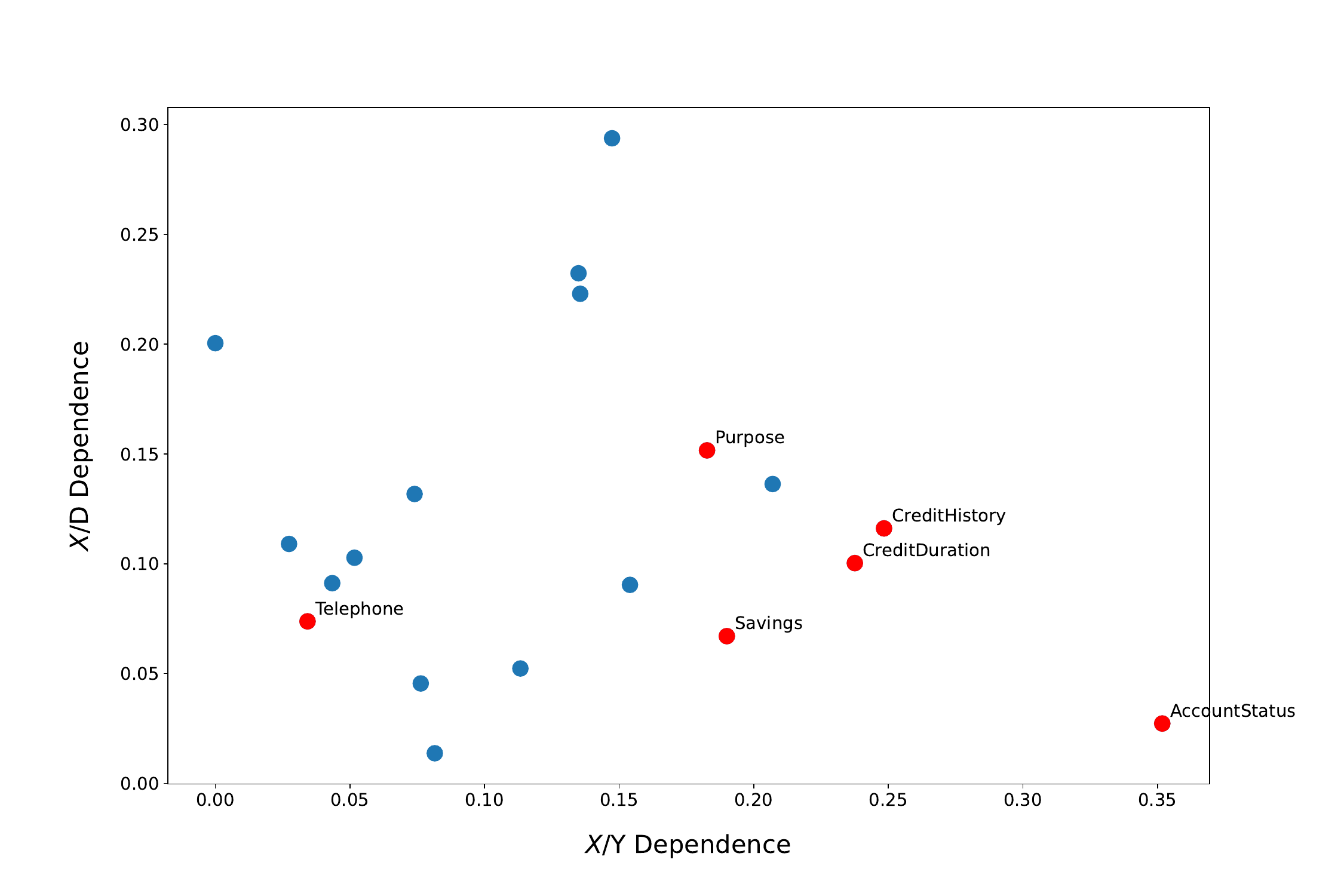} 
		\vskip 0.3cm 
		\label{Figure_VCramer_TREEs} 
	\end{figure}

	\noindent In Figure \ref{Figure_VCramer_TREEs}, we present a scatter plot of Cramer's V, which is a measure of association between two variables.\footnote{The Cramer's V varies from 0 (corresponding to no association between the variables) to 1 (complete association).} The $X$-axis displays the association between each explanatory variable and the target variable whereas the $Y$-axis does so for each explanatory variable and gender. By construction, the top-right region of the plot includes variables that are both gendered and useful to classify good and bad-type borrowers. Such variables would be natural candidates to generate a gender effect since (1) they are likely to be selected by the algorithm because of their strong correlation with the target and (2) they proxy for gender. However, our dataset does not include such variables, yet (as we will show below) several of our credit scoring models will exhibit a lack of fairness. 
 
	\subsection{Credit scoring models}
	
	\noindent We place ourselves in the configuration of a bank that seeks to assess the creditworthiness of some loan applicants through the development of a credit scoring model. Before modeling credit default, we apply various pre-treatments to the dataset. First, we transform the 11 categorical variables into binary variables. Doing so is standard practice as it gives more degrees of freedom to the algorithms and permits to better exploit their inherent flexibility. Second, we exclude the binary variable \textit{foreign worker} from the explanatory variables to solely evaluate the algorithms' fairness with respect to gender, avoiding the inclusion of another protected attribute. Such pre-treatment leads us with a total of 55 explanatory variables. We do not split the dataset into a training sample and a test sample, but estimate all scoring models using the whole sample to have enough observations when conducting conditional fairness tests.\footnote{In Section \ref{section:discussion}, we use a larger dataset and properly distinguish between the training and the test samples.}\medskip
	
	\noindent We estimate a set of credit scoring models to predict default, namely logistic regression (LR), classification tree (TREE), random forests (RF), XGBoost (XGB), support vector machine (SVM), and artificial neural networks (ANN). Thus, we consider both standard parametric regression models (LR) and machine-learning models, able to extract non-linear relationships. We estimate individual classifiers (SVM, TREE, ANN) as well as ensemble methods (RF, XGB), which have proved to perform well in credit scoring applications (\cite{lessmann2015benchmarking}). Furthermore, our analysis includes natively interpretable models or white-box models (LR, TREE) and black-box models (RF, ANN) as our fairness diagnosis method can accommodate both types of models. In practice, the performance of ML\ models (and, as shown below, their fairness diagnosis) is quite sensitive to the value of the parameters used to fine tune the model and control the learning process. In the present study, we determine hyperparameter values using a ten-fold cross-validation and a random search algorithm.\footnote{We split the dataset into ten subsamples. We use nine of them for in-sample calibration, while using the remaining one for out-of-sample testing. This procedure is carried out ten times by changing the subsample used out-of-sample. Within this process, the random search algorithm trains the considered credit scoring models based on different hyperparameter settings. Finally, the random search algorithm chooses the hyperparameter values with the highest average accuracy across all subsamples. See Table \ref{Appendix_HyperparameterA} in Appendix for more information about hyperparameters.} \medskip
 
    \noindent In a first step, we train the credit scoring models by including gender in the feature space and refer to these models as the \textit{with-models}.\footnote{To ensure that gender was actually used in the machine learning model, such as the RF, we rely on a feature importance methodology called \textit{permutation importance}. In this approach, the feature importance is determined by evaluating the decrease in the model's performance when the values of this variable are randomly reshuffled. In our situation, if there is a drop in the model's AUC after reshuffling the gender variable, this means that gender has been used in the model. We verify that the permutation importance associated with gender on the AUC of the model is different from zero for all of the considered ML models.} While this may not be a particularly realistic setting, as banking regulation typically prevents lenders from including gender in their scoring models, it constitutes a useful reference point in our controlled experiment. Panel A of Table \ref{Table1} displays the performance of the six \textit{with-models}: we measure statistical performance using the percentage of correct classification (PCC), the area under the curve (AUC), and the false discovery rate (FDR). We see that all considered models perform well as the PCC\ ranges between $76.4$ and $87.3$, the AUC between $0.811$ and $0.938$, and the FDR between $0.1369$ and $0.1799$. To assess the economic performance of the model, we follow \cite{lessmann2015benchmarking} and compute a misclassification cost defined as the weighted sum of the false positive rate (FPR) and the false negative rate (FNR). The corresponding cost function is expressed as $CF=C_{+|-}\times FPR+C_{-|+}\times FNR$, where $C_{+|-}$ denotes the cost of granting credit to a bad-type borrower, and $C_{-|+}$ the opportunity cost that results from denying credit to a good-type borrower. We set $C_{+|-}$ to two and $C_{-|+}$ to one to reflect the fact that an actual default is more costly for the bank than the rejection of a good-type borrower. As shown in Table \ref{Table1}, the best models identified by the AUC and CF criteria are not the same. For instance, out of the six models, XGB is the second-best according to the AUC but the second-last according to CF.\medskip
	
	\noindent In a second step, we re-train all models after removing gender from the feature space. We call these models the \textit{without-models}. In this case, if a model exhibits a lack of fairness, it necessarily originates from a single or a set of features acting as proxy for gender. Panel B of Table \ref{Table1} displays the PCC, AUC, FDR, and CF values for the six \textit{without-models}. Overall, the performance levels are comparable with those of the \textit{with-models}, as none of the performance measures changes by more than six percentage points.\footnote{As we only consider \textit{with-models} that selects the gender variable, we disregard some models with a higher in-sample performance. As a result, some in-sample performance metrics are higher without gender than with gender.}
	
	
	\subsection{Fairness diagnosis}
 
	\label{Section_Fairness_Diagnostic}

	\noindent We now implement the statistical tests outlined in Section \ref{Section_Fairness_Inference} to assess the fairness of the algorithms. We first consider the six alternative fairness measures described in Section \ref{Section Fairness Metrics} and the six \textit{with-models}. For each model, we report in Panel A of Table \ref{Table2} the p-value associated with a given null hypothesis of fairness.\footnote{The test for conditional statistical parity is based on two classes obtained using a K-Prototype clustering algorithm (cf. Figure \ref{fig:Appendix_Risk_Groups} in the Appendix).} Overall, the message is clear as the null hypothesis is, as expected, rejected at the 95\% confidence level for most model - fairness definition combinations.\medskip
	
	
	\noindent We then investigate whether the scoring models also lead to unfairness when gender is not used in the analysis. In Panel B, we display similar results as in Panel A but for \textit{without-models}. For most models, the null hypothesis is no longer rejected, indicating that the previously detected unfairness stemmed from including gender.\footnote{This result may seem surprising given that the impossibility theorem of \cite{Kleinberg2017InherentTI} states that no more than one of the three fairness metrics of statistical parity, predictive parity and equal odds can hold at the same time for a well calibrated classifier, a same data generating process, and a protected attribute. However, the impossibility theorem concerns the true joint distribution of $(Y,\widehat{Y},D)$ that we do not observe in practice. When using a finite sample of loans and accounting for estimation risk, it is possible to not reject the null for two or more fairness definitions. Inference here allows us to consider the uncertainty associated with the estimates, while controlling for the risk of falsely rejecting the fairness null hypothesis.} Differently, with the ANN model, we reject the null fairness hypothesis as the p-values associated with (conditional) statistical parity are around $0.01$. For this model, removing the gender variable is not a sufficient condition to safeguard members of the protected group.\medskip
	 
	 \noindent In an additional step, we illustrate the important role played by operational risk. While it is well known that small changes in ML model hyperparameters can lead to large variations in performances (see \cite{bergstra_random_nodate}), here we focus on another type of operational risk. Indeed, we show that the choice of the parameters used to control the learning process of credit scoring models can have strong consequences in terms of fairness. To illustrate our point, we consider an alternative TREE\ model, called TREE-prime, which is based on a slightly modified set of hyperparameters. Both the TREE and TREE-prime models rely on a procedure commonly used in the credit scoring industry, based on a $k$-fold cross-validation and a random search algorithm (\cite{lessmann2015benchmarking}). The only difference between the two models lies in the set of values considered for the random search of the optimal hyperparameters.\footnote{\label{hyper_PRIME}In the TREE-prime model, we reduce the set of values for the maximum depth of the tree from 1-29 to 1-9, we increase the set of values for the minimum number of instances required to split a node from 2-9 to 2-59, and we increase the set of values for the minimum number of individuals by leaf from 1-19 to 1-59.} This alternative model complies with current model design and training rules and could be implemented in production by a lender. The TREE-prime model leads to a slightly lower performance (e.g., PCC = $79.0$ vs. $81.5$, AUC = $0.8393$ vs. $0.8866$). However, we see in Table \ref{Table4} that the new model leads to drastically different conclusions: we reject the null hypothesis of fairness for virtually all metrics at the 95\% confidence level.\medskip
 
    \noindent We generalize our results on the effect of hyperparameter values on fairness in the context of a larger-scale experiment. We slightly modify the hyperparameter grid of the TREE model, i.e., the maximum depth of the tree (Max depth), the minimum number of individuals required to split an internal node (Min sample split) or to be at a leaf node (Min sample leaf). Specifically, the Max depth goes from 10 to 60 by increments of 5 and the Min sample split and Min sample leaf range from 6 to 50 with a step size of 5. This configuration results in 891 distinct hyperparameter grids, each corresponding to a unique decision tree model. Among these models, we only kept those for which we reject the null hypothesis of fairness according to statistical parity at the 95\% confidence level and the AUC ranges from 0.83 to 0.92. We ended up with 33 decision tree models which are all comparable to TREE-prime. We report in Table \ref{table:other_primes} in the Appendix the optimal hyperparameters, the performance criteria, and the p-value associated with the fairness test for 10 models (randomly selected among the 33).
	
	\subsection{Interpretability}
	
	\label{Section Fairness Intepretability}
	
	\noindent In the previous section, we have identified several scoring models that are classified as unfair by our testing procedure. We are now going to identify the variables at the origin of this lack of fairness. As an example, we consider the TREE-prime model as it has been shown to exhibit a severe lack of fairness.\medskip
	
\begin{figure}[!htbp] 
		\centering 
		\caption{Fairness PDP for the statistical parity in TREE-prime model} 
		\includegraphics[width=1.10\textwidth,trim={4.5cm 6cm 0 6cm},clip]{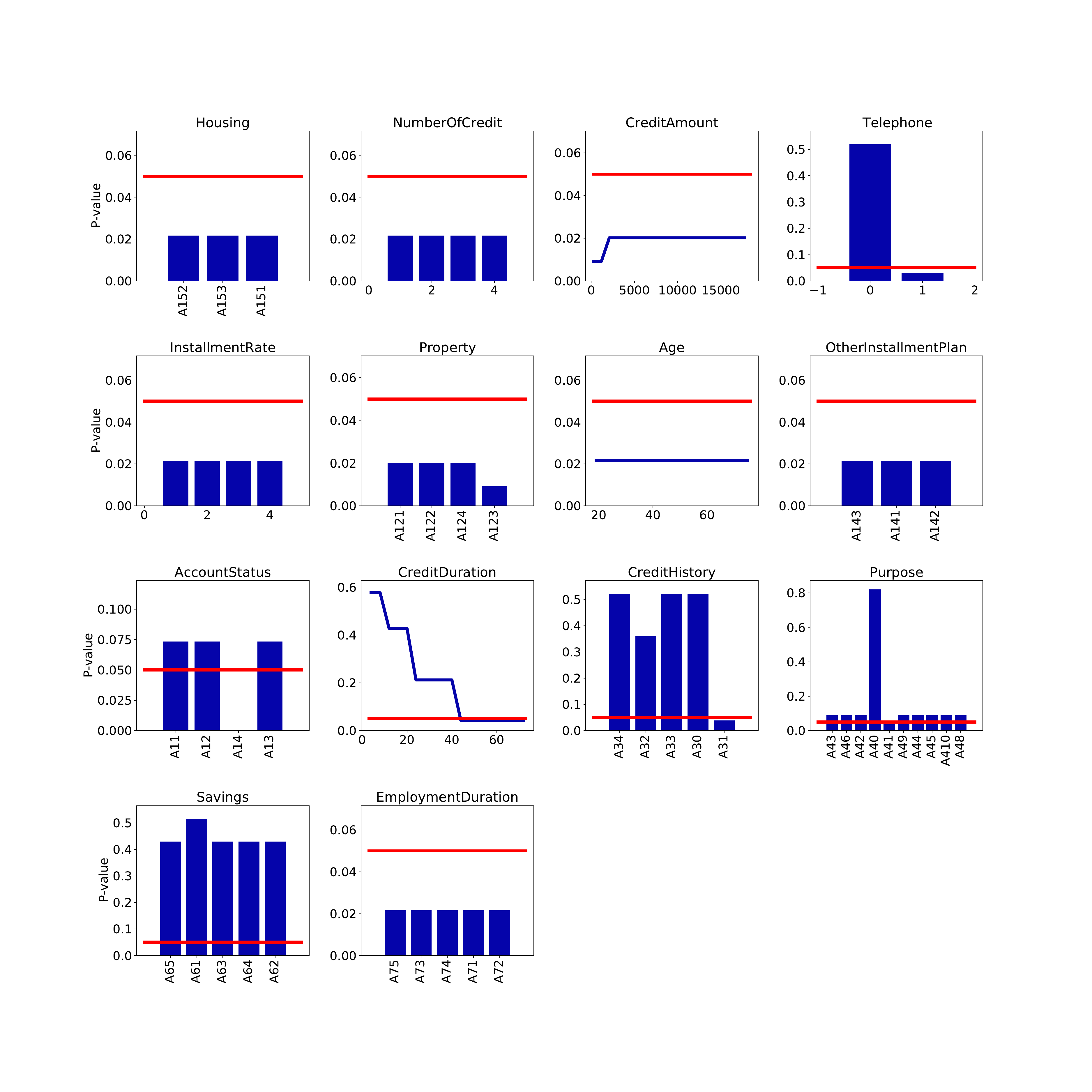} 
		\vskip 0.3cm 
		\label{Figure_FPDP_TREE_Statistical Parity} 
	\end{figure}

	\noindent We start by generating the FPDP associated with the statistical-parity null hypothesis. The individual plots associated with the 14 features used in the model are reported in Figure \ref{Figure_FPDP_TREE_Statistical Parity}. Recall that in the initial sample (see Table \ref{Table4}), the statistical parity test leads to the rejection of the null hypothesis of fairness (i.e., p-value<5\%). As explained in the previous section, a feature is said to be a candidate variable if setting the same value to all applicants leads to not rejecting the null hypothesis anymore (i.e., p-value>5\%). In Figure \ref{Figure_FPDP_TREE_Statistical Parity}, we identify six candidate variables, namely \textit{credit duration}, \textit{credit history}, \textit{purpose}, \textit{savings}, \textit{account status}, and \textit{telephone}.\footnote{For the \textit{account status} variable, the test statistic cannot be defined for one category since all predicted outcomes are equal to 1, meaning that it is independent from gender and that the null hypothesis is not rejected.} Interestingly, we show in Figures \ref{fig:FPDP_CP}, \ref{fig:FPDP_EO} and \ref{fig:FPDP_EOP} in the Appendix that we identify the same six candidate variables when using, as an alternative, conditional statistical parity, equal odds, or equal opportunity.\medskip 
	
	\noindent When contrasting the six candidate variables with the decision tree in Figure \ref{Figure_Fairness_Decision} in the Appendix, we note the following. First, all the candidate variables are features that partition the data space into leaves associated with positive (blue) and negative (orange) labels. Second, all the features used to split the applicants between positive and negative labels are not necessarily identified as candidate variables (e.g., property). Collectively, these observations confirm the soundness of the FPDP method.\medskip 
	
	
	\noindent In Figure \ref{Figure_VCramer_TREEs}, we display for all features their dependence (Cramer's V) with the target variable and with gender. We highlight in red the six features identified as candidate variables. Five of them are significantly correlated with the target variable and/or with gender. These five variables appear legitimate in lending and are regularly used in credit scoring. Differently, the remaining candidate variable, \textit{Telephone}, appears to be much less correlated with the target and gender variables.

    \subsection{Mitigation}

    \label{application_mitigation}
	\noindent We now proceed with the mitigation of the lack of fairness using the methods presented in Section \ref{section:Mitigation}. We display the results in Table \ref{tab:Mitigation} for the two competing approaches: (1) re-estimation in Panel A and (2) neutralization of a feature in Panel B. In each case, we report the p-values associated with the various fairness definitions, as well as the predictive performance measures (AUC, PCC, FDR, and CF). We find two important results when re-estimating the model with one less variable. First, as shown in Panel A, removing the variable and re-estimating the model does not always mitigate the lack of fairness. For instance, removing \textit{Credit History} (or \textit{Purpose} or \textit{Savings}) leads to rejecting the null hypothesis for statistical parity. This persistent fairness problem is akin to the instability issue of ML models discussed by \cite{mullainathan_machine_2017}. Indeed, removing one feature from the model and retraining it generally leads to select another feature or a combination of features which are correlated with the removed one. Second, when re-estimating the models actually permits the mitigation of unfairness, it almost invariably results in a substantial decline in accuracy. For instance, the AUC drops from 0.8393 to 0.6727 when discarding \textit{Account Status} and to 0.8017 when removing \textit{Credit Duration}.\medskip

 
	\noindent Given these shortcomings, we suggest not re-estimating the model. Indeed, Panel B shows that setting the value of some candidate variables not only solves the fairness problem (p-value>5\% for all fairness definitions) but alters only marginally the performance. When one aims to maximize the model's performance, the most favorable model is achieved by setting \textit{Telephone} to 0 for all applicants. In this case, we never reject the fairness hypothesis and the AUC only drops from 0.8393 to 0.8325. Interestingly, fixing the value of the other candidate variables also solves the fairness problem but the resulting drop in performance is stronger (e.g. \textit{Account Status}, \textit{Credit History}, or \textit{Credit Duration}). This result makes perfect sense as these three variables display the strongest positive dependence with the target variable, hence very much help the model to deliver high predictive performance.\medskip

\begin{figure}[!h] 
    \centering 
    \caption{Accuracy-fairness trade-off} 
    \bigskip
    \includegraphics[width=0.9\textwidth,trim={0cm 0cm 0 0cm},clip]{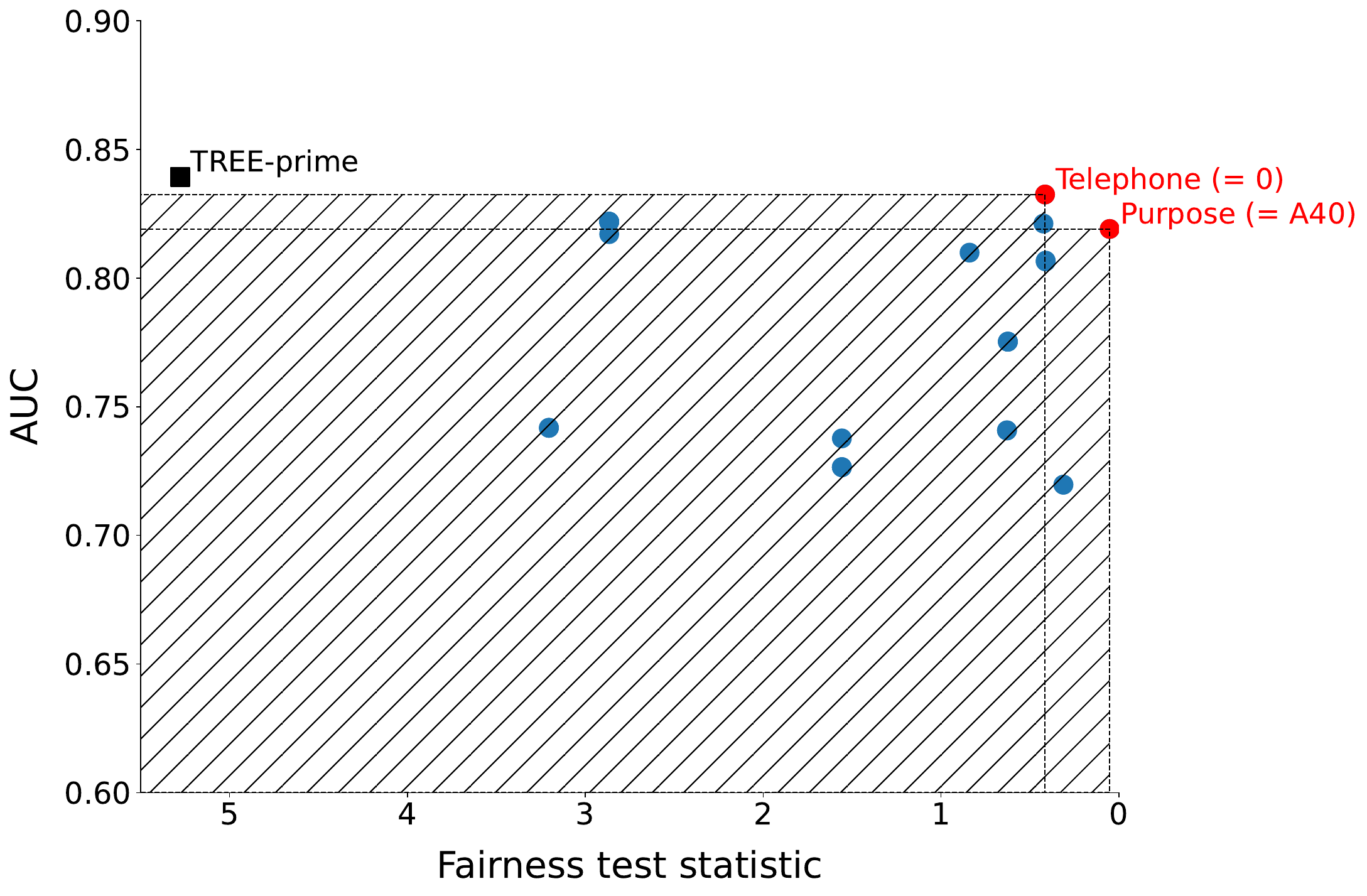} 
    \label{Figure 4} 

    \justifying
    \noindent {\footnotesize Notes: This figure displays the relationship between the AUC and the fairness test statistic associated with the null hypothesis of statistical parity, for the modified TREE-prime models and after mitigation. Each point represents a modified model linked to a candidate variable and a specific value. The values correspond to those reported in Panel B of Table \ref{tab:Mitigation}. Red dots correspond to Pareto optimal models that dominate the others, both in terms of performance and fairness. A model dominates another one in terms of fairness if its test statistic is lower than the test of the other model.}
\end{figure}

    \noindent As shown in Section \ref{section:Mitigation}, an alternative mitigation approach is to 
    rely on the value the fairness test statistics. In this case, our dual goal is to increase both the performance and the degree of fairness. In Figure \ref{Figure 4}, we display the AUC and the fairness test statistic (statistical parity) associated with the models for which we have "muted" a candidate variable (see Table \ref{tab:Mitigation}). The Pareto frontier, depicted in red, is formed by the two Pareto-optimal mitigation strategies, dominating the others both in terms of performance and fairness. The first strategy entails neutralizing the impact of the variable \textit{Telephone} by setting its value to 0 (none) for all individuals. The resulting model displays an AUC of $0.8325$ and a p-value of $0.5195$ (vs. $\mbox{AUC}=0.8393$ and $\mbox{p-value}=0.0216$ for the original TREE-prime model, represented by a black square). The second involves neutralizing the influence of the variable \textit{Purpose} by assigning its value to the modality A40.\medskip

 
    \noindent Interestingly, when considering the economic cost instead of the statistical accuracy, we only obtain one Pareto optimal solution, both minimising the economic cost and maximising the fairness. As shown in Figure \ref{Figure Fairness vs Costs German} in the Appendix, we find that the model obtained by setting the variable \textit{Purpose} at the modality A40 is optimal both in terms of costs and fairness metrics. Although we get the same optimal model as that found for the AUC/fairness trade-off, this is not a universal conclusion. We notice that the model obtained by fixing \textit{Telephone} at 0 is no longer optimal in terms of economic costs. This illustrates the divergence between economic and accuracy evaluations \citep{lessmann2015benchmarking} and the usefulness of cost-sensitive approaches \citep{Petrides2022}.\medskip 


    \noindent Finally, we evaluate the "marginal rate of substitution", which indicates the extent to which lenders must compromise their profits to achieve fairness. In Figure \ref{Figure cost ratios German} in the Appendix, we depict the average misclassification costs calculated for all mitigation strategies (based on the 12 candidate variables/values reported in Table \ref{tab:Mitigation}) when $C_{+|-}$ varies from $2$ to $50$.\footnote{In contrast to \cite{lessmann2015benchmarking}, we do not adjust the probability threshold of the scoring model when $C_{+|-}$ varies. In fact, we refrain from modifying the model and threshold to ensure that the baseline model (TREE-prime) remains unfair and the model after mitigation consistently passes the fairness test. Adjusting the threshold to minimize cost may not always produce a fair model, even if it's often the case.} These average costs have been normalized to represent the percentage increase compared to the misclassification associated with the original TREE-prime model, the model for which we invalidated the null hypothesis of fairness. Hence, a positive value signifies an enhanced cost to achieve fairness. Our findings indicate that achieving fairness necessitates an average cost increase of between $11.5\%$ to $14.5\%$. This average "marginal rate of substitution" underscores the potential economic trade-off associated with fairness. However, this average value masks significant variations across mitigation methods. In some instances, achieving fairness can even lead to a cost reduction. This is exemplified by the optimal model obtained by setting the variable Purpose to modality A40. As shown in Table \ref{tab:Mitigation}, this model reduces misclassification costs to $1.0819$ compared to $1.1852$ for the initial TREE-prime model, representing a decrease of $8.71\%$. \medskip

    \noindent A similar Pareto front approach has been recently used by \cite{Kozodoi2023} to compare different fairness definitions and mitigation approaches in a profit-oriented credit scoring context using real-world data. Their results are in line with ours, as they show that reducing discrimination to a reasonable extent is possible while maintaining a relatively high profit.\medskip

    \noindent Mitigating unfairness can also have a significant impact on lenders' actions. When one removes a candidate feature to address a lack of fairness, it unavoidably modifies credit-granting decisions. For example, in a simple logistic regression model, muting a candidate variable by fixing its value is tantamount to altering the intercept. This, in turn, affects the predicted probability for all individuals. If the probability threshold employed by the bank to grant loans remains unchanged, it will necessarily modify the number of loans granted.\medskip
    
    \noindent To maintain a consistent number of approved loans before and after mitigation, a natural approach involves adjusting the probability threshold used by the bank. Regardless of the model employed (linear logistic regression or nonlinear ML models), adjusting the probability threshold influences the fairness statistic, accuracy measures (except AUC), and misclassification costs. However, we demonstrate in Section \ref{section:loansgranted} in the Appendix that this adjustment does not impair the effectiveness of our mitigation method. 
    
\section{External validity and discussion} 

\label{section:discussion}


\noindent To confirm the external validity of our approach, we replicate our entire framework using an alternative dataset, namely the Taiwan credit card dataset \citep{UCI}. This dataset includes information on demographic factors (e.g., gender, age, marital status), credit history, and bill statements of $30,000$ credit card clients in Taiwan between April 2005 and September 2005. Among these clients, $60\%$ are female. Besides the 23 features, we also know whether each client defaulted on his or her payment: out of the $30,000$ clients, $6,636$ or $22.1\%$ of them, defaulted. Before applying our fairness framework, we conducted data cleaning, which led to retaining $25,983$ clients in the database with 13 features. Note that we changed the encoding of the target variable such that "1" refers to non-default and "0" to default, as in the German credit dataset.\medskip 

\noindent Using a larger database allows us to show that our framework works on a sample distinct from the training sample. The interest in applying our methodology to the test sample stems from the fact that the models are subject to overfitting on the training data, leading us to overestimate its performance and potentially biasing the results of our statistical tests.\footnote{An alternative would be to apply the "honest" approach proposed by \cite{Athey_2016}. However, as shown in Table \ref{honest_estimation} in the Appendix, the honest approach and the standard train/test split procedure result in an identical fairness diagnosis and comparable predictive performances (AUC, PCC, FDR). Hence, we present solely the outcomes of the standard procedure.} Therefore, we split our remaining $25,983$ observations into a training ($67\%$) and a test sample ($33\%$). We end up with $17,408$ and $8,575$ observations in the training and test samples. All of reported results were obtained on the test dataset.

\noindent Applying our framework to the Taiwan credit card dataset provides the following takeaways. First, we do not reject the null hypothesis of fairness for most of the models and fairness definitions. As reported in Table \ref{Table 3 Taiwan} in the Appendix, we only reject the null hypothesis of fairness according to statistical parity for the XGB and the ANN models at the 95\% confidence level. We also reject the null hypothesis of fairness for the XGB model according to equal opportunity. Our inability to reject the null hypothesis of fairness should not be attributed to insufficient statistical power, given the large size of our dataset (30,000 observations). Instead, it suggests that gender does not have a significant correlation with individual defaults. Regarding the performances of the models, as shown in Table \ref{tab:perf_taiwan} in the Appendix, we see that all considered models perform well on the test sample as the PCC ranges between 78.3 and 81.4, the AUC between 0.7335 and 0.7711, and the FDR between 0.1427 and 0.1731. 

\medskip

\noindent Second, as shown in Figure \ref{FPDP Taiwan} in the Appendix, we detect four candidate variables: the \textit{Credit history}, the \textit{Bill statement}, the \textit{Funding amount} and the \textit{Previous payment}. 
 Third, we mitigate the lack of fairness by assigning the value of 3,082 to \textit{Previous payment} for all individuals. As shown in Table \ref{tab:Mitigation_Taiwan} in the Appendix, the performance of the resulting model is very close to the original XGB (AUC=0.7674 vs. AUC=0.7701). This confirms the efficiency of our mitigation method. 
Fourth, as shown in Figure \ref{Figure 4 Taiwan} in the Appendix, we find four Pareto optimal XGB models, dominating the others both in terms of AUC and fairness. For these four models located on the Pareto frontier, we identify two candidates variables. We can either neutralize the variable \textit{Previous payment} by setting its value to 3,082 or 12,328, or neutralize the variable \textit{Bill statement} by assigning its value to 53,196 or 17,732. This result contrasts with the German Credit dataset, where only a single Pareto optimal model exists, eliminating the necessity to select among multiple optimal solutions. 


\medskip

\noindent Finally, we identify different Pareto optimal XGB models when we consider the statistical accuracy (Figure \ref{Figure 4 Taiwan} in the Appendix) or the economic cost (Figure \ref{Figure Fairness vs Costs Taiwan} in the Appendix). When considering statistical accuracy, we obtain the four Pareto optimal XGB models mentioned in the previous paragraph. However, when focusing on economic cost, as shown in Figure \ref{Figure Fairness vs Costs Taiwan} in the Appendix, we find three Pareto optimal XGB models. More importantly, among these models, only one is optimal both in terms of statistical accuracy and economic cost. It corresponds to the XGB model obtained by setting the value of \textit{Bill Statement} to 53,196. This modified model achieves an AUC of 0.7608, a CF of 1.3708, and a p-value for the fairness test statistic of 0.5786, while the analogous metrics for the initial XGB model before mitigation (represented by a black square on Figure \ref{Figure 4 Taiwan} and \ref{Figure Fairness vs Costs Taiwan} in the Appendix) amounted to 0.7701, 1.33, and 0.0218, respectively. 


\section{Conclusion}
	
\noindent Credit scoring algorithms can be life-changing for many households and businesses. Indeed, they dictate the eligibility for credit and the terms under which it is granted. Therefore, ensuring these algorithms adhere to fair lending practices is crucial, particularly when they utilize sophisticated, opaque machine learning techniques and extensive datasets. In this paper, we propose a framework designed to formally test the null hypothesis of fairness and enable lenders and regulatory bodies to identify the factors contributing to unfair outcomes. A key advantage of our interpretability method is to give us many degrees of freedom to mitigate the fairness issue, allowing for adjustments based on both statistical and economic criteria.
\medskip
	
\noindent The high legal and regulatory uncertainties surrounding the use of ML algorithms acts as an impediment for financial service providers to innovate and invest in screening technologies \citep{evans_keeping_2017,bartlett_algorithmic_2020}. Furthermore, recent ruling in Europe against companies using discriminatory algorithms \citep{geiger_court_nodate} as well as debate in the US to potentially ban some automated tools used by corporations \citep{givens_opinion_2021} put a spotlight on this concerns. As algorithmic discrimination of women and minorities may be completely unintentional and can be embedded in data and/or in the inner workings of algorithms, it is more important than ever for lenders and their regulators to benefit from clear guidelines and tools able to red flag any form of group discrimination. We believe our methodology can contribute to provide such guidelines and regulatory tools.
\medskip
	
\noindent While we focus on access to credit, our methodology also can prove useful for studying other life-changing decision-making algorithms. Indeed, similar algorithms are in use in the fields of money laundering and fraud detection, automated claim management in the insurance industry, predictive justice, automated screening of job applicants, or university admission.

\clearpage

\baselineskip=0.6\normalbaselineskip

\bibliographystyle{apalike}
\bibliography{main.bib}

    \clearpage 

    \noindent {\footnotesize Notes: This figure displays the dependence of each feature with respect to the target (horizontal axis) and the gender (vertical axis), using Cramer's V as a measure of dependence. Red dots correspond to candidate variables according to the statistical parity test applied to the TREE-prime model (see Section \ref{Section Fairness Intepretability}).}

     \newpage
	
	\noindent {\footnotesize Notes: Each subplot displays the FPDP for statistical parity, associated with a given feature and the classification TREE-prime model. The Y-axis displays the p-value of the statistical parity test statistic. The red line represents the 5\% threshold.}

	\newpage

\clearpage

  \begin{sidewaystable}[h] 
 \centering 
 \caption{Literature review} 
 \label{Table 1 Litterature Review} 
 \vspace{2mm} 
 
 \resizebox{1\textwidth}{!}{ 
 \renewcommand{\arraystretch}{1.2}
\begin{tabular}{l|ccc|c|c|c|c|c|c}
\toprule
\multirow{2}{*}{Reference} & \multicolumn{3}{c|}{Datasets} & \multirow{2}{*}{Protected Attributes}  & \multirow{2}{*}{Credit Scoring} &  \multirow{2}{*}{Criteria} &  \multirow{2}{*}{Inference}  &  \multirow{2}{*}{Mitigation} & \multirow{2}{*}{Model-agnostic}  \\ \cmidrule{2-4}  
              & Nb. & Obs. & Var. & & &  & & & \multicolumn{1}{c}{}     \\ \midrule 
\textcolor{blue}{\cite{bartlett_algorithmic_2020}}     & 1 & 800 & 3 & Gender & &  IND  & \checkmark{} & PRE   &   \\ 
\cite{CaldersKamiran2009} & 1 & 1,000 & 20 & Age/Foreigners & \checkmark{} & IND   & & PRE         & \checkmark{}  \\ 
\cite{Calders2009} & 2 & 24,921 & 17 & Gender & \checkmark{} & IND  & & PRE          & \checkmark{}   \\ 
\cite{kamiran2012data} & 3 & 80,187 & 50 & Gender/Race & & IND  & \checkmark &  PRE          & \checkmark{}  \\ 
\cite{xu2018fairgan} & 1 & 48,842 & 14 & Gender & & IND & &   PRE       &  \checkmark{} \\ 
\cite{adel2019one} & 2 & 18,919 & 13 & Gender/Race & & IND + SEP   & \checkmark &   PRE       & \checkmark{}  \\ 
\cite{Kairouz2022} & 4 & 17,703 & 30,207 & Gender/Marital Status/Identity & & IND + SEP  & &  PRE       & \checkmark{}  \\
\cite{petrovic2022fair} & 4 & 32,261 & 274 & Gender/Race/Age & \checkmark{} & IND + SEP  & &   PRE       & \checkmark{}  \\
\cite{jiang2020identifying} & 5 & 27,011  & 37 & Gender/Race & \checkmark{} & IND + SEP  & &   PRE       & \checkmark{}  \\
\cite{Madras2018} & 2 & 54,421 & N/A & Gender/Age & & IND + SEP & &   PRE       & \checkmark{}  \\ 
\cite{xu2019achieving}  & 1 & 65,123 & 11 & Gender & & CAU & &   PRE      & \checkmark{}  \\ 
\cite{Yang2023}          & 5 & 14,584 & 25 & Race &  &   SEP &   & IN   &    \\ 
\cite{pope2011implementing}          & 1 & 590,924 & 9 & Gender/Race/Age &  &   IND & \checkmark{}  & IN   &    \\ 
\cite{Nabi2018}                      & 2 & 29,921 & 13 & Gender/Race & & CAU & & IN         &  \checkmark{} \\  
\cite{Nabi2019}                      & 1 & 5,278 & 6 & Race & & CAU   & & IN         &  \checkmark{} \\  
\cite{Kamiran2010}                      & 4 & 85,083 & 40 & Gender/Race & &  IND  & & IN + POST    &   \\ 
\cite{calders2010three} & 1 & 48,842 & 14 & Gender & & IND & &  IN + POST &   \\ 
\cite{Hardt2016} & 1 & 301,536 & - & Race & \checkmark{} & SEP  & & POST  & \checkmark{}  \\ 
\cite{Chiappa_2019}                  & 2 & 24,621 & 17 & Gender & \checkmark{} & CAU  & & POST  & \checkmark{}   \\ 
 \cite{Kilbertus2017} & - & - & - & - & & CAU & &   POST     & \checkmark{}  \\ 
 \cite{pleiss2017fairness}            & 3 & 20,249 & 13 & Gender/Race/Age & &  SEP + SUF  & & POST  &   \checkmark{}        \\ 
\cite{noriega2019active} & 4 & 30,263 & 59 & Age/Race/Rural areas & \checkmark{} & SEP + SUF   & & POST       & \checkmark{}    \\ 
  \cite{Krishna2018} & - & - & - & - & & IND + SEP  & &  POST   & \checkmark{}  \\ 
\cite{Valera2018} & 1 & 11,000 & 12 & Race & & IND + SEP & &  POST   & \checkmark{}  \\ 
\cite{kamiran2012decision}           & 2 & 9,138 & 68 & Gender/Race & & IND + SEP  & & POST   &  \checkmark{} \\
\cite{iosifidis2019fae} & 2 & 42,590 & 16 & Gender/Marital Status  & & SEP & &  PRE + POST  &   \\ 
\cite{Kozodoi2023} & 7 & 80,738 & 88 & Age  & \checkmark{} & IND + SEP + SUF &  & PRE + IN + POST  &   \\
 \cite{Galhotra2017} & 2 & 23,000 & 17 & Gender/Race &      \checkmark{} & IND + CAU & \checkmark &  & \checkmark{}    \\ 
\cite{kim2023fair} & 1 & 129,457 & 3,123 & Gender/Age/Nb. Children/etc. & \checkmark{} & IND + SEP + SUF  & &         & \checkmark{}   \\ 
\cite{Liu2019} & 2 & 28,028 & 10 & Gender/Race & & IND + SEP + SUF &  &     & \checkmark{}   \\ 
\textbf{Our paper} & \textbf{2} & \textbf{13,491} & \textbf{13} & \textbf{Gender} & \textbf{\checkmark{}} & \textbf{IND + SEP + SUF}  & \textbf{\checkmark{}} & \textbf{POST}  & \textbf{\checkmark{}}   \\ 
\bottomrule
\end{tabular}
}

\medskip \justifying
	\noindent {\footnotesize Notes: In columns 2, 3, and 4, we report the number of datasets used in each paper, along with the corresponding number of observations and variables linked to those datasets. When there is more than one dataset, we report the average number of observations and variables. Columns 5, 7, and 9 display the protected attribute, the fairness criteria, and the mitigation method used in each paper. Columns 6, 8, and 10 indicate whether the paper includes an empirical application using a credit scoring dataset, whether it uses or develops inference, or whether its mitigation method is model-agnostic.
\noindent\newline
IND: independence, SEP: separation, SUF: sufficiency, CAU: causal, PRE: pre-processing, IN: in-processing, POST: post-processing.}
 \end{sidewaystable}

 \clearpage
 
 \begin{table}[H] 
		\centering 
		\caption{Model performances with and without the protected feature} 
		\label{Table1} 
		\vspace{0.5mm}
    
			\begin{tabular}{c c c c c c c}
				\toprule 
                \addlinespace[10pt]
                \multicolumn{7}{c}{Panel A: Models with gender} \vspace{7pt} \\ \midrule
				{} & LR  & TREE & RF & XGB & SVM & ANN \\
				\midrule 
				AUC & 0.8279 & 0.8266 & 0.9380 & 0.8877 & 0.8107 & 0.8341 \\
                PCC & 77.4 & 77.3 & 87.3 & 81.3 & 78.2 & 79.1 \\
                FDR & 0.1643 & 0.1626 & 0.1369 & 0.179 & 0.1717 & 0.1799 \\
                CF & 0.9305 & 0.9214 & 0.7471 & 1.0162 & 0.9714 & 1.0214 \\  
                \toprule \addlinespace[10pt]
                \multicolumn{7}{c}{Panel B: Models without gender} \vspace{7pt} \\ \midrule
				{} & LR  & TREE & RF & XGB & SVM & ANN \\
				\midrule 
				AUC & 0.8264 & 0.8866 & 0.9372 & 0.8261 & 0.8059 & 0.8754 \\
                PCC & 77.2 & 81.5 & 87.4 & 79.6 & 76.0 & 81.1 \\
                FDR & 0.1629 & 0.1358 & 0.1358 & 0.1861 & 0.1788 & 0.1669 \\
                CF & 0.9229 & 0.7671 & 0.7405 & 1.0614 & 1.0133 & 0.9405 \\
				\bottomrule 
		\end{tabular} 
  
  \medskip \justifying \noindent {\footnotesize Notes: This table reports the area under the ROC curve (AUC), the percentage of correct classification (PCC), the False Discovery Rate (FDR), and a Cost Function (CF) values for each scoring model, with gender (Panel A) and without gender (Panel B). The CF is defined as a weighted average of type-I and type-II errors, where the weight associated with the type-I (type-II) error is equal to 2 (1). LR: Logistic Regression, TREE: classification tree, RF: Random Forest, XGB: XGBoost, SVM: Support Vector Machine, ANN: Artificial Neural Network.}
	\end{table}

\begin{table}[h]
\centering
\caption{Fairness tests for models with and without gender}

\label{tab:performance_german} 
		
\vspace{0.5mm}
  
 \vspace{2mm}\label{Table2} 
        \begin{tabular}{p{3.2cm}p{1.1cm}p{1.1cm}p{1.1cm}p{1.1cm}p{1.1cm}p{1.1cm}}\toprule 
        \addlinespace[10pt]
        \multicolumn{7}{c}{Panel A: Models with gender} \vspace{7pt}\\ \midrule
        {} &       \multicolumn{1}{c}{LR} &     \multicolumn{1}{c}{TREE} &      \multicolumn{1}{c}{RF}  &   \multicolumn{1}{c}{XGB}    &  \multicolumn{1}{c}{SVM}     &  \multicolumn{1}{c}{ANN}     \\
        \midrule
        Statistical parity    &  0.0003* &  0.0097* &  0.0349* &  0.0000* &  0.0041* &  0.0041* \\
        Cond. parity &  0.0003* &  0.0110* &  0.0434* &  0.0000* &  0.0022* &  0.0017* \\
        Equal odds            &  0.0185* &   0.2387 &   0.8220 &  0.0004* &   0.1436 &   0.0802 \\
        Equal opportunity     &   0.0888 &   0.3012 &   0.7796 &  0.0004* &   0.1675 &   0.6554 \\
        Predictive equality   &  0.0242* &   0.1801 &   0.5753 &   0.0945 &   0.1598 &  0.0277* \\
        Sufficiency &   0.8990 &  0.0128* &   0.9284 &   0.6024 &   0.8790 &   0.8128 \\ 
        \toprule
        \addlinespace[10pt]
        \multicolumn{7}{c}{Panel B: Models without gender} \vspace{7pt}\\ \midrule
        {} &       \multicolumn{1}{c}{LR} &     \multicolumn{1}{c}{TREE} &      \multicolumn{1}{c}{RF}  &   \multicolumn{1}{c}{XGB}    &  \multicolumn{1}{c}{SVM}     &  \multicolumn{1}{c}{ANN}     \\
        \midrule
        Statistical parity    &  0.0734 &  0.5310 &  0.1206 &   0.0965 &   0.2913 &  0.0067* \\
        Cond. parity &  0.0590 &  0.3998 &  0.1542 &   0.0841 &   0.3821 &  0.0048* \\
        Equal odds            &  0.6712 &  0.5645 &  0.9242 &   0.7202 &   0.6754 &   0.1727 \\
        Equal opportunity     &  0.7746 &  0.8892 &  0.7796 &   0.4213 &   0.5175 &   0.6602 \\
        Predictive equality   &  0.3977 &  0.2890 &  0.7783 &   0.9216 &   0.5451 &   0.0685 \\
        Sufficiency &  0.6068 &  0.3572 &  0.7772 &   0.3943 &   0.1787 &   0.4920 \\\bottomrule 
        \end{tabular}

        \medskip \justifying
        \noindent {\footnotesize Notes: This table reports the p-values of the fairness tests (see Section \ref{Section_Fairness_Inference}) obtained for six different scoring models, with the gender variable (Panel A) and without the gender variable (Panel B). * indicates statistical significance at 5\%. LR: Logistic Regression, TREE: classification tree, RF: Random Forest, XGB: XGBoost, SVM: Support Vector Machine, ANN: Artificial Neural Network.}

\end{table}

\begin{table}[h] 
 \centering 
 \caption{Fairness tests for the TREE models} 
 \label{Table4} 
\begin{tabular}{l l l}
 \toprule 
{} & \multicolumn{1}{c}{TREE} & \multicolumn{1}{c}{TREE-prime} \\
\midrule 
Statistical parity & 0.5310 & \hspace{0.3cm} 0.0216* \\
Cond. parity & 0.3998 & \hspace{0.3cm} 0.0153* \\
Equal odds & 0.5645 & \hspace{0.3cm} 0.0363* \\
Equal opportunity & 0.8892 & \hspace{0.3cm} 0.0101* \\
Predictive equality & 0.2890 & \hspace{0.3cm} 0.8852 \\
Sufficiency & 0.3572 & \hspace{0.3cm} 0.0934 \\
\midrule 
AUC & 0.8866 & \hspace{0.3cm} 0.8393 \\
PCC & \hspace{0.1cm} 81.5 & \hspace{0.65cm}79.0 \\
FDR  & 0.1358 & \hspace{0.3cm} 0.2041 \\
CF & 0.7671 & \hspace{0.3cm} 1.1852 \\
\bottomrule 
\end{tabular} 

\medskip \justifying
\noindent {\footnotesize Notes: This table reports the p-values of the fairness tests obtained for two different decision trees. The first one (TREE) corresponds to the decision tree without the gender variable. TREE-prime denotes a decision tree obtained with the same feature space, but with an alternative hyperparameter tuning strategy. * indicates statistical significance at 5\%.}
 \end{table}
 
	\clearpage
	\begin{table}[h] 
 \centering 
 \caption{Mitigation} 
 \label{Table 1} 
 \vspace{2mm} 

\begin{tabular}{p{4.1cm}lcccc}
\toprule
\addlinespace[10pt]
\multicolumn{6}{c}{Panel A: With re-estimation} 
\vspace{7pt}\\ \midrule
{} &  \multicolumn{1}{c}{SP}  &     AUC &   PCC &  FDR &    CF  \\
\midrule
TREE-prime      &              0.0216*  &  0.8393 &  79.0 &    0.2041 &  1.1852 \\
\midrule
Telephone      &              0.1019 &  0.8380 &  78.7 &    0.2041 &  1.1843 \\
CreditHistory  &              0.0462* &  0.8336 &  78.2 &    0.2061 &   1.1967 \\
CreditDuration &              0.6463 &  0.8017 &  75.5 &    0.2153 &   1.2510 \\
Purpose        &              0.0263* &  0.7804 &  74.6 &    0.2170 &   1.2586 \\
Savings        &              0.0332* &  0.7130 &  72.7 &    0.2615 &  1.6157 \\
AccountStatus  &              0.9875 &  0.6727 &  71.8 &    0.2743 &  1.7333  \\
\toprule 
\addlinespace[10pt]
\multicolumn{6}{c}{Panel B: Without re-estimation} \vspace{7pt} \\ \midrule
TREE-prime      &              0.0216* &  0.8393 &  79.0  &    0.2041  &  1.1852 \\
\midrule
Telephone (= 0) & 0.5195 & 0.8325 & 77.8 & 0.1912 & 1.0924 \\
Purpose (= A49) & 0.0905 & 0.8219 & 76.8 & 0.2181 & 1.2795  \\
Savings (= A61) & 0.5150 & 0.8212 & 77.0 & 0.2106  & 1.2243  \\
Purpose (= A40) & 0.8206 & 0.8191 & 76.7 & 0.1899  & 1.0819   \\
Purpose (= A43) & 0.0905 & 0.8171 & 76.8 & 0.2181 & 1.2795  \\
CreditHistory (= A32) & 0.3596 & 0.8099 & 75.6 & 0.2143 & 1.2443  \\
CreditHistory (= A34) & 0.5212 & 0.8068 & 77.7 & 0.2311  & 1.3924  \\
Savings (= A65) & 0.4296 & 0.7753 & 73.9 & 0.2346 & 1.3890  \\
AccountStatus (= A13) & 0.0734 & 0.7418 & 72.9  & 0.2066 & 1.1781  \\
CreditDuration (= 20.0) & 0.4277 & 0.7408 & 72.6 & 0.2715 & 1.7167 \\
CreditDuration (= 24.0) & 0.2120 & 0.7377 & 68.2 & 0.2264 & 1.2819 \\
CreditDuration (= 8.0) & 0.5767 & 0.7197 & 71.2 & 0.2854 & 1.8467 \\
\bottomrule
\end{tabular}

\label{tab:Mitigation}

\medskip \justifying
\noindent {\footnotesize Notes: This table reports the mitigation results obtained with re-estimation (Panel A) and without re-estimation (Panel B) of the TREE-prime model. Panel A displays the p-values of the fairness test according to statistical parity (SP) and the performance metrics (AUC, PCC, FDR, CF) obtained after removing one candidate variable from the list of explanatory variables. In this case, the  model is re-estimated. Panel B displays the p-values of the fairness test according to statistical parity (SP) and the performance metrics (AUC, PCC, FDR, CF) obtained after setting one candidate variable to the same value for all individuals. The value of the candidate variable is reported in parentheses.}
 
 \end{table}	

\appendix
	
\renewcommand{\thefigure}{A\arabic{figure}}
	
\setcounter{figure}{0}
	
\renewcommand{\thetable}{A\arabic{table}}
	
\setcounter{table}{0}

\makeatletter
    \renewcommand\thesection{}
    \renewcommand\thesubsection{A.\@arabic\c@subsection}
    \def\@seccntformat#1{\csname #1ignore\expandafter\endcsname\csname the#1\endcsname\quad}
    \let\sectionignore\@gobbletwo
    \let\latex@numberline\numberline
    \def\numberline#1{\if\relax#1\relax\else\latex@numberline{#1}\fi}
    \makeatother
    
	\setcounter{table}{0}

	\section{Online Appendix of "The Fairness of Credit Scoring models"}

    \medskip
    \medskip
    
	\subsection{Proof of theorem Fairness test}
	
	\label{Appendix_Fairness_Test}

    \begin{proof} 
    The likelihood ratio is defined as:
    \begin{align}
        \Lambda=\frac{\prod_{k=1}^{K} \prod_{u=0}^{1} \prod_{v=0}^{1} p_{u,v,k}(\hat{\theta}_k)^{n_{u,v,k}}}
        {\prod_{k=1}^{K} \prod_{u=0}^{1} \prod_{v=0}^{1} \hat{p}_{u,v,k}^{n_{u,v,k}}}
    \end{align}
    with $\hat{p}_{u,v,k}=n_{u,v,k}/n_k$, $p_{u,v,k}(\hat{\theta}_k)=\hat{\alpha}_{u,k} \hat{\beta}_{v,k}$, $\hat{\alpha}_{u,k}=\sum_{v=0}^{1}n_{u,v,k}/n_k$, and $\hat{\beta}_{v,k}=\sum_{u=0}^{1}n_{u,v,k}/n_k$. The Likelihood-Ratio test statistic of $H_{0,i}$ satisfies: 
    \begin{align}
        \begin{split}
        F_{H_{0,i}} & =-2\log(\Lambda) \\
        & = 2n \sum_{k=1}^{K} \sum_{u=0}^{1} \sum_{v=0}^{1} \hat{p}_{u,v,k} 
        \log \left( \frac{\hat{p}_{u,v,k}}{p_{u,v,k}(\hat{\theta}_k)} \right) \\
        & = 2n \sum_{k=1}^{K} \sum_{u=0}^{1} \sum_{v=0}^{1} 
        (\hat{p}_{u,v,k}-p_{u,v,k}(\hat{\theta}_k)+p_{u,v,k}(\hat{\theta}_k) )
        \log \left( 1 + \frac{\hat{p}_{u,v,k}-p_{u,v,k}(\hat{\theta}_k)}{p_{u,v,k}(\hat{\theta}_k)} \right)
        \end{split}
    \end{align}
    If $H_{0,i}$ is true, then $\hat{p}_k$ and $p(\hat{\theta}_k)$ will both tend to the same limit in probability for $n_k$ sufficiently large, so that $\hat{p}_k - p(\hat{\theta}_k)$ will be of small order. Denote $y_i=((\hat{p}_{u,v,k}-p_{u,v,k}(\hat{\theta}_k))/p_{u,v,k}(\hat{\theta}_k))$. Under the above-mentioned assumptions, the Maclaurin series expansion of the natural logarithm function around $0$ is given by $\log(1 + y_i) = y_i - y_i^2/2 + y_i^3/3 \dots$ as $|y_i| < 1$ and $y_i$ converges to 0 in probability. Hence,
    \begin{align}
        \begin{split}
        F_{H_{0,i}}  = 2n \sum_{k=1}^{K} \sum_{u=0}^{1} \sum_{v=0}^{1}  &
         \Big\{
          (\hat{p}_{u,v,k}-p_{u,v,k}(\hat{\theta}_k))
        +\frac{(\hat{p}_{u,v,k}-p_{u,v,k}(\hat{\theta}_k))^2}{p_{u,v,k}(\hat{\theta}_k)} \\
       & -\frac{(\hat{p}_{u,v,k}-p_{u,v,k}(\hat{\theta}_k))^2}{2p_{u,v,k}(\hat{\theta}_k)} 
        +O_p(\hat{p}_{u,v,k}-p_{u,v,k}(\hat{\theta}_k))^3
        \Big\} 
        \end{split}    
    \end{align}
    As $\sum_{u=0}^{1} \sum_{v=0}^{1}(\hat{p}_{u,v,k}-p_{u,v,k}(\hat{\theta}_k))=0$, the LR test statistic can be approximated as follows:
    \begin{align}
        F_{H_{0,i}} \approx 2n  \sum_{k=1}^{K} \sum_{u=0}^{1} \sum_{v=0}^{1} 
        \frac{(\hat{p}_{u,v,k}-p_{u,v,k}(\hat{\theta}_k))^2}{2p_{u,v,k}(\hat{\theta}_k)} 
        =\sum_{k=1}^{K} \sum_{u=0}^{1} \sum_{v=0}^{1}
        \frac{(n_{u,v,k}-n_kp_{u,v,k}(\hat{\theta}_k))^2}{n_kp_{u,v,k}(\hat{\theta}_k)}
    \end{align} 
    When $H_{0,i}$ is true, the LR test statistic $F_{H_{0,i}}$ is asymptotically distributed as a Chi-square with $q$ degrees of freedom, i.e., the number of free parameters for the model under the null hypothesis.
    
    \end{proof}
    
	\clearpage
 
	\subsection{Database description}
	
	\label{Appendix_Database_Description}
	
	\begin{table}[h]
		\caption{Database description}
		\label{TabX}
		\vspace{-3mm}
		\begin{center}
			\begin{tabular}{cccc}
				\toprule
				Short name & Complete name & Variable type & Domain \\ \midrule
				Age\smallskip & Age & Numerical\smallskip & $%
				\mathbb{R}
				^{+}$ \\ 
				CreditAmount & Credit amount & Numerical\smallskip & $%
				\mathbb{R}
				^{+}$ \\ 
				CreditDuration & Credit duration & Numerical\smallskip & $%
				\mathbb{R}
				^{+}$ \\ 
				AccountStatus & Status of existing checking account & Categorical\smallskip
				& $\#4$ \\ 
				CreditHistory & Credit history & Categorical\smallskip & $\#5$ \\ 
				Purpose & Credit Purpose & Categorical\smallskip & $\#10$ \\ 
				Savings & Status of savings accounts and bonds & Categorical\smallskip & $%
				\#5 $ \\ 
				EmploymentDuration & Employment length & Categorical\smallskip & $\#5$ \\ 
				InstallmentRate & Installment rate & Numerical\smallskip & $\{1,2,3,4\}$ \\ 
				Gender\&PersonalStatus & Personal status and gender & Categorical$%
				_{\smallskip }$ & $\#4$ \\ 
				Guarantor & Other debtors & Categorical\smallskip & $\#3$ \\ 
				ResidenceTime & Period of present residency & Numerical\smallskip & $%
				\{1,2,3,4\}$ \\ 
				Property & Property & Categorical\smallskip & $\#4$ \\ 
				OtherInstallmentPlan & Installment plans & Categorical\smallskip & $\#3$ \\ 
				Housing & Residence & Categorical\smallskip & $\#3$ \\ 
				NumberOfCredit & Number of existing credits & Numerical\smallskip & $%
				\{1,2,3,4\}$ \\ 
				Job & Employment & Categorical\smallskip & $\#4$ \\ 
				NumberLiablePeople & Dependents & Numerical\smallskip & $\{1,2\}$ \\ 
				Telephone & Telephone & Binary\smallskip & $\#2$ \\ 
				ForeignWorker & Foreign worker & Binary\smallskip & $\#2$ \\ 
				CreditRisk & Credit score & Binary\smallskip & $\#2$ \\ \bottomrule
			\end{tabular}%
		\end{center}
		\par
	\end{table}
	\newpage

	\begin{table}[h!]
		\caption{Feature overview}
		\label{feature_overview}
		\vspace{-0.5cm}
		\begin{center}
			\resizebox{0.85\textwidth}{!}{
			\begin{tabular}{cc}
				\toprule\toprule
				Complete name & Description \\ \midrule \midrule
				Age & Age in years \\ \midrule
				Credit amount & Credit amount \\ \midrule
				Credit duration & Duration in month \\ \midrule
				Status of existing checking account & 
				\begin{tabular}{c}
					A11 : ... < 0 DM, \ A12 : 0 <= ... <
					200 DM \\ 
					A13 : $\geq 200$ DM / salary assignments (1 year) \\ 
					A14 : no checking account%
				\end{tabular}
				\\ \midrule
				Credit history & 
				\begin{tabular}{c}
					A30: no credits taken/ all credits paid back duly \\ 
					A31: all credits at this bank paid back duly \\ 
					A32: existing credits paid back duly till now \\ 
					A33: delay in paying off in the past \\ 
					A34: other credits existing (not at this bank)%
				\end{tabular}
				\\ \midrule
				Credit Purpose & 
				\begin{tabular}{c}
					A40: car (new), A41: car (used), A42: equipment \\ 
					A43: radio/television, A44: domestic appliances \\ 
					A45: repairs, A46: education, A48: retraining \\ 
					A49: business, A410: others%
				\end{tabular}
				\\ \midrule
				Status of savings accounts and bonds & 
				\begin{tabular}{c}
					A61: $<100$ DM, A62: $100\leqslant $ $x<500$ \\ 
					A63: $500\leqslant $ $x<1000,$ A64: $\geq 1000$ DM \\ 
					A65: unknown/ no savings account%
				\end{tabular}
				\\ \midrule
				Employment duration & 
				\begin{tabular}{c}
					A71: unemployed, A72: $.<1$ year A73: $1\leqslant $ $x<4$ years, \\ 
					A74 : $4\leqslant $ $x<7$ years, A75: $\geq 7$ years%
				\end{tabular}
				\\ \midrule
				Installment rate & Installment rate in percentage of disposable income \\ 
				\midrule
				Personal status and gender & 
				\begin{tabular}{c}
					A91: male : divorced/separated \\ 
					A92: female : divorced/separated/married \\ 
					A93: male : single, A94: male : married/widowed%
				\end{tabular}
				\\ \midrule
				Other debtors & A101: none, A102: co-applicant \\ \midrule
				Period of present residency & Present residence since \\ \midrule
				Property & 
				\begin{tabular}{c}
					A121: real estate, A123: car or other, \\ 
					A122: building society savings agreement \\ 
					A124: unknown / no property%
				\end{tabular}
				\\ \midrule
				Installment plans & A141: bank, A142: stores, A143: none \\ \midrule
				Housing & A151: rent, A152: own, A153: for free \\ \midrule
				Number of existing credits & Number of existing credits at this bank \\ 
				\midrule
				Employment & 
				\begin{tabular}{c}
					A171: unemployed/ unskilled - non-resident \\ 
					A172: unskilled - resident A173: skilled employee \\ 
					A174: management/ self-employed/highly qualified%
				\end{tabular}
				\\ \midrule
				Dependents & Number of people being liable to provide maintenance \\ \midrule
				Telephone & A191: none, A192: yes \\ \midrule
				Foreign worker & A201: yes, A202: no \\ \midrule
				Credit score & 1: Good, 2: Bad \\ \bottomrule
			\end{tabular}}%
		\end{center}
		\par 
	\end{table}

    \clearpage

	\subsection{Hyperparameters of machine-learning models}
	
	\label{Appendix_hyperparameter_tuning}
	
	\begin{table}[h!]
		\caption{Hyperparameter tuning}
		\label{Appendix_HyperparameterA}\medskip 
		\vspace{-1.5mm}
		\resizebox{1\textwidth}{!}{
			\begin{tabular}{lcccccc}
				\toprule 
				\textbf{Hyperparameter set}   & \multicolumn{1}{c}{TREE}  & TREE-prime  & ANN & SVM & RF & XGBoost \\ \hline 
				& & & & & & \\
				\multicolumn{1}{l}{Criterion \smallskip}  & \smallskip Entropy, Gini (Gini) & Entropy, Gini (Gini) &  &   & Entropy, Gini (Entropy) & \\ 
				\multicolumn{1}{l}{Max. depth of the tree \smallskip} \smallskip  & 1-29 (20) & 1-9 (7) &    & & 1-99 (82) & 1-5 (2) \\ 
				\multicolumn{1}{l}{Min. number of individuals required to split a node
					\smallskip} \smallskip  & 2-9 (2) & 2-59 (56) &    & & 2-49 (3) &\\ 
				\multicolumn{1}{l}{Min. number of individuals by leaf \smallskip} \smallskip  & 1-19 (5)
				& 1-59 (18) & &  &  1-29 (10)&\\ 
				\multicolumn{1}{l}{Number of inputs compared at each split \smallskip} \smallskip & 
				all, $\sqrt{k},\log 2(k)$ ($\sqrt{k}$) & all, $\sqrt{k},\log 2(k)$ (all) & &  & all, $\sqrt{k},\log 2(k)$ (all) &\\ 
				\multicolumn{1}{l}{Decrease of the impurity greater than or equal to this value \smallskip}\smallskip  & 0-0.9 (0) & 0-0.9 (0) &  &  & 0-0.9 (0)  &\\ 
				\multicolumn{1}{l}{Optimization method: Grid Search \smallskip} \smallskip  & & &  
				\multicolumn{1}{c}{Yes} & & & \\ 
				\multicolumn{1}{l}{Number of hidden layers  \smallskip} \smallskip  & & & 
				\multicolumn{1}{c}{1}& & &\\ 
				\multicolumn{1}{l}{Number of neurons by hidden layer  \smallskip}\smallskip & & &\multicolumn{1}{c}{1-25 (20)}&&& \\ 
				\multicolumn{1}{l}{Max. number of iterations \smallskip}\smallskip & & &\multicolumn{1}{c}{250}&&& \\ 
				\multicolumn{1}{l}{Activation function \smallskip}\smallskip  & & &\multicolumn{1}{c}{relu}&&& \\ 
				\multicolumn{1}{l}{Solver for weight optimization \smallskip}\smallskip & & &\multicolumn{1}{c}{adam}&&& \\ 
				\multicolumn{1}{l}{L2 penalty (regularization term) parameter \smallskip}\smallskip  & & &\multicolumn{1}{c}{0.0001}&&& \\ 
				\multicolumn{1}{l}{Early stopping \smallskip}\smallskip & & &\multicolumn{1}{c}{True}&&& \\ 
				\multicolumn{1}{l}{Max. number of epochs to not meet tol improvement \smallskip}\smallskip & & &\multicolumn{1}{c}{50}&&& \\ 
				\multicolumn{1}{l}{Regularization parameter\smallskip} \smallskip  & & & &  1 & &\\
				\multicolumn{1}{l}{Kernel\smallskip} \smallskip   & &  & &  linear & & \\
				\multicolumn{1}{l}{Learning rate\smallskip} \smallskip &  & & & & & 0-0.9 (0.5) \\ 
				\multicolumn{1}{l}{Min. sum of instance weight (hessian) needed in a child\smallskip \smallskip}   &  & &  & & & 1-39 (15) \\ 
				\multicolumn{1}{l}{Min. loss reduction required to make a further partition of a leaf node}\smallskip \smallskip  & & & &  & & 0-0.9 (0.3)\\ 
				\multicolumn{1}{l}{Subsample ratio of columns when constructing each tree\smallskip \smallskip}  & & & &  & & 0-0.9 (0.4)\\ 
				\multicolumn{1}{l}{Subsample ratio of columns for each level\smallskip \smallskip}  & & & & &  & 0-0.9 (0.1)\\ 
				\multicolumn{1}{l}{Subsample\medskip\ ratio of columns for each node (split)\smallskip \smallskip}  & & & & &  & 0-0.9 (0.2)\\ 
				\bottomrule
			\end{tabular}%
		}
  
        \medskip
        
		\noindent \justifying {\footnotesize Notes: This table displays the hyperparameter tuning
		procedures for the various machine-learning models used in the numerical
		analysis: Decision Tree (Tree), Artificial Neural Network (ANN), Support
		Vector Machine (SVM), Random Forest (RF), and XGBoost. The values in
		parentheses are the optimal hyperparameter values.}
	\end{table}
 
    \clearpage

    \subsection{Alternative TREE-prime models}
	
	\label{Appendix_operational_risk}

     \begin{table}[h]
    \centering
    \caption{Alternative TREE-prime models}
    \vspace{2mm} 
    \label{table:other_primes}
    \resizebox{\textwidth}{!}{ 
    \begin{tabular}{ccccccclcccc}
    \toprule
    \multirow{2}{*}{Index} & Min. sample &  Min. sample &  Min. impurity & Max.  &  Max. & \multirow{2}{*}{Criterion} &  \multicolumn{1}{c}{\multirow{2}{*}{SP}} &       \multirow{2}{*}{AUC} &    \multirow{2}{*}{PCC} &      \multirow{2}{*}{FDR} &       \multirow{2}{*}{CF} \\ 
    & split & leaf & decrease & features &  depth &  &   &    &   &   &  \\
    \midrule
    TREE & 2 & 5 & 0 & sqrt& 20 & gini & 0.5310 & 0.8866 & 81.5 & 0.1358 & 0.7671   \\
    1 & 8 & 5 & 0 & sqrt & 13 & gini & 0.0041* & 0.8856 & 82.1 & 0.1604 & 0.9000 \\
    2 & 9 & 12 & 0 & max. & 8 & gini & 0.0092* & 0.8847 & 81.2 & 0.1614 & 0.9076 \\
    3 & 28 & 8 & 0 & max. & 8 & gini & 0.0079* & 0.8835 & 81.6 & 0.1551 & 0.8705 \\
    4 & 2 & 5 & 0 & log2 & 11 & gini & 0.0006* & 0.8721 & 80.4 & 0.1519 & 0.8562 \\
    5 & 26 & 3 & 0 & sqrt & 14 & entropy & 0.0046* & 0.8645 & 80.6 & 0.1773 & 1.0052 \\
    6 & 4 & 5 & 0 & sqrt & 8 & entropy & 0.0025* & 0.8601 & 79.3 & 0.1513 & 0.8562 \\
    7 & 56 & 18 & 0 & max. & 7 & gini & 0.0216* & 0.8393 & 79.0 & 0.2041 & 1.1852  \\
    8 & 45 & 12 & 0 & max. & 6 & entropy & 0.0458* & 0.8359 & 77.8 & 0.2127 & 1.2443 \\
    9 & 15 & 6 & 0 & log2 & 12 & gini & 0.0145* & 0.8342 & 77.9 & 0.1965 & 1.1276 \\
    10 & 35 & 1 & 0 & log2 & 11 & gini & 0.0334* & 0.8330 & 77.8 & 0.1839 & 1.0452 \\
    \bottomrule
    \end{tabular}}
    
    \medskip \justifying
    \noindent {\footnotesize Notes: This table displays the optimal hyperparameters, performances (AUC, PCC, FDR, CF), and fairness test statistics according to statistical parity (SP) for alternative TREE-prime models. The hyperparameters correspond to the optimal one obtained after 10-fold cross-validation when changing the hyperparameter grid between models. The first row of the table refers to the TREE model mentioned in column 1 of Table \ref{Table4}.}
    \end{table}
    
	\clearpage
    \subsection{Feature distribution by gender}

    \begin{figure}[!htbp] 
		\centering 
		\caption{Feature distributions} 
		\includegraphics[width=1.10\textwidth,trim={5cm 5.5cm 0 6cm},clip]{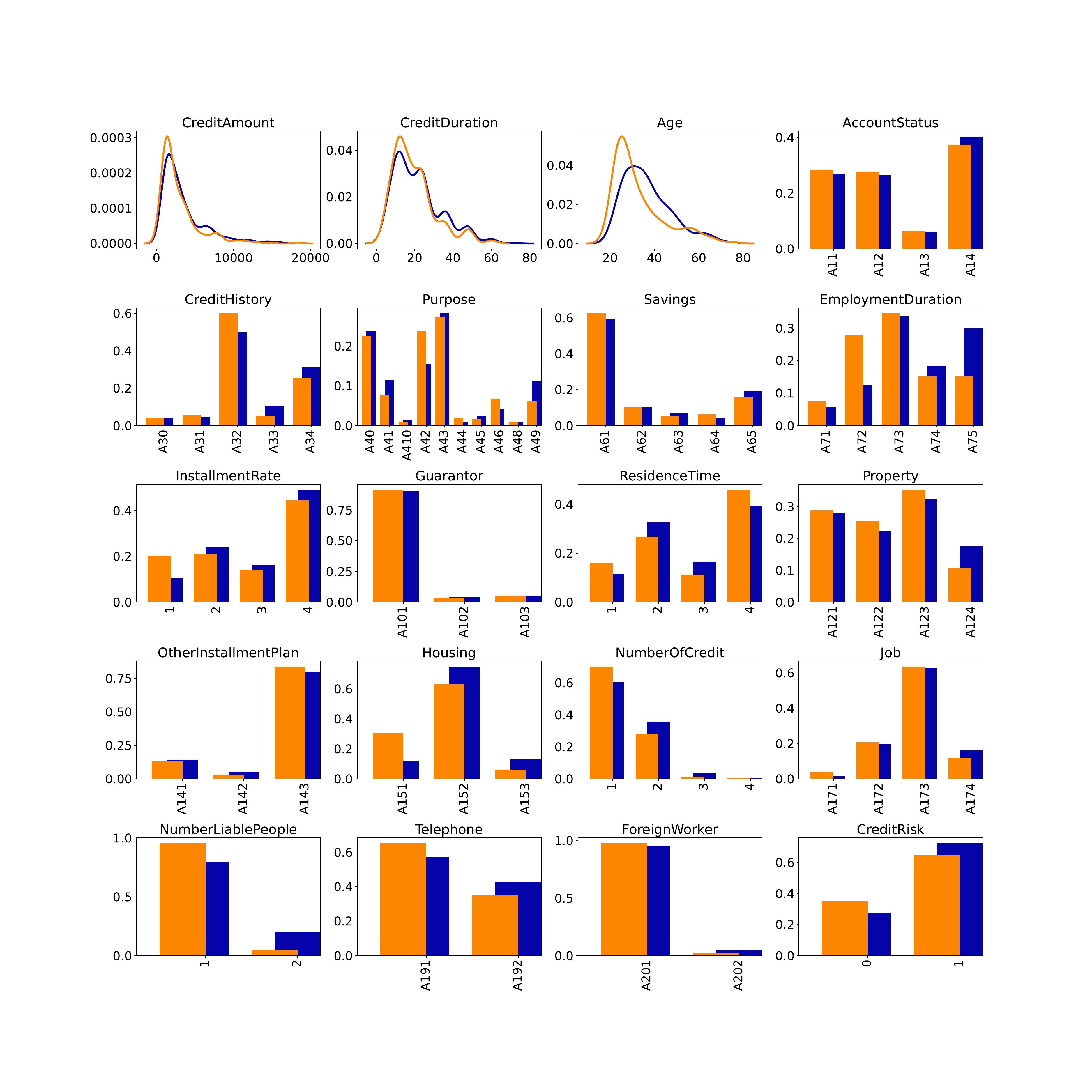} 
		\vskip 0.3cm 
		\label{%
			Figure1 Features Distributions} 
	\end{figure}
	\noindent {\footnotesize Notes: These figures display the feature distributions by
		gender, using kernel density estimation for continuous variables. Blue color
		refers to men and orange to women. }

    \clearpage

	\subsection{Feature distribution by class of risk}
	
	\label{section:Appendix_Risk_Groups}
	
	\begin{figure}[!htbp] 
		\centering 
		\caption{Feature distribution by class of risk} 
		\includegraphics[width=1.10\textwidth,trim={5cm 5.5cm 0 6cm},clip]{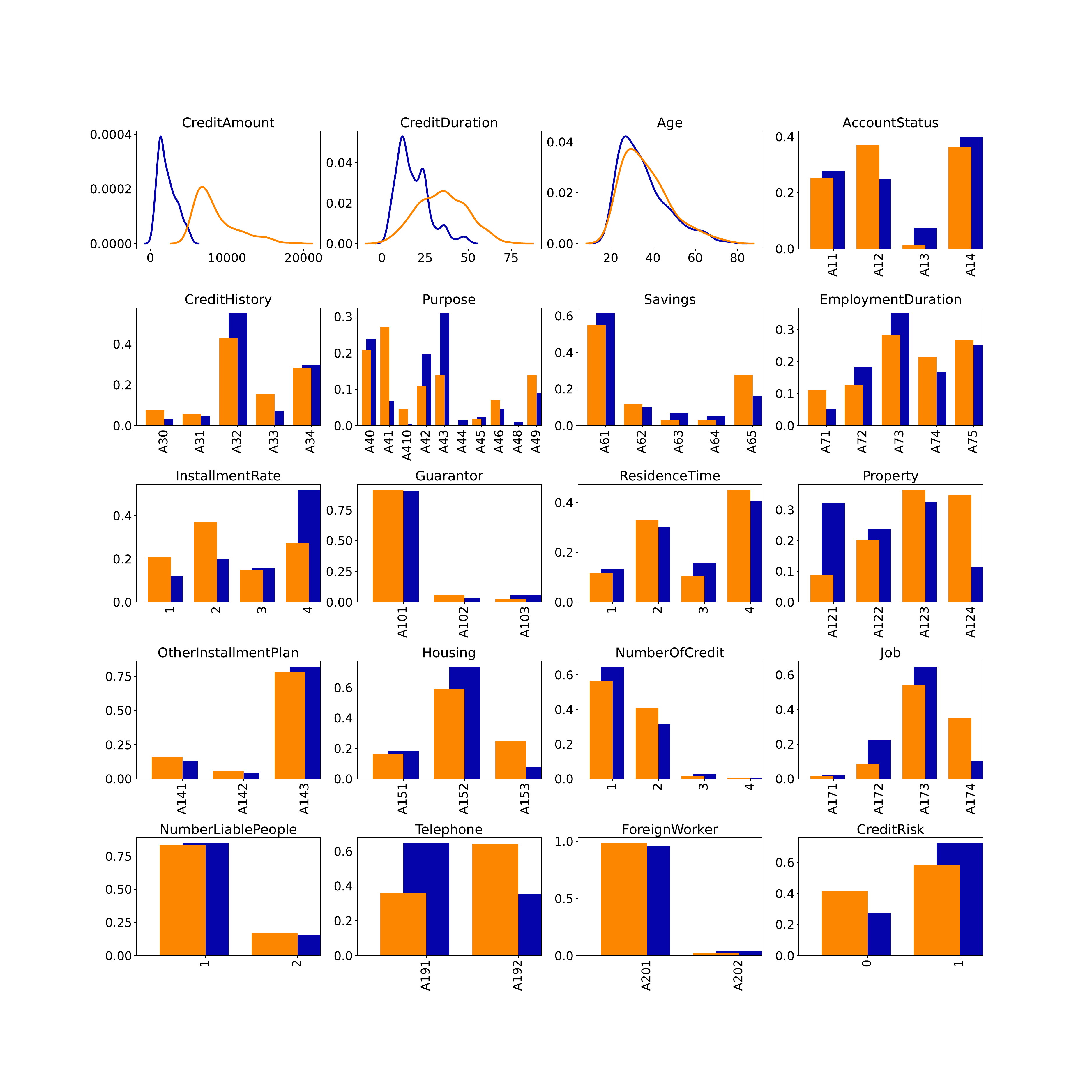} 
		\label{fig:Appendix_Risk_Groups} 
	\end{figure}
	\vspace{-0.8cm}
	{\noindent \footnotesize Notes: These figures display the feature distributions by class of risk, using kernel density estimation for continuous variables. Blue color refers to Class 1 (low-risk profiles) and orange to Class 2 (high-risk profiles).}
	\clearpage 
	
	\subsection{Decision tree}
	\begin{figure}[!htbp] 
		\centering 
		\caption{Decision tree for the TREE-prime model} 
		\includegraphics[width=1.2\textwidth,height=1\textwidth,trim={8cm 7.5cm 0 7cm},clip]{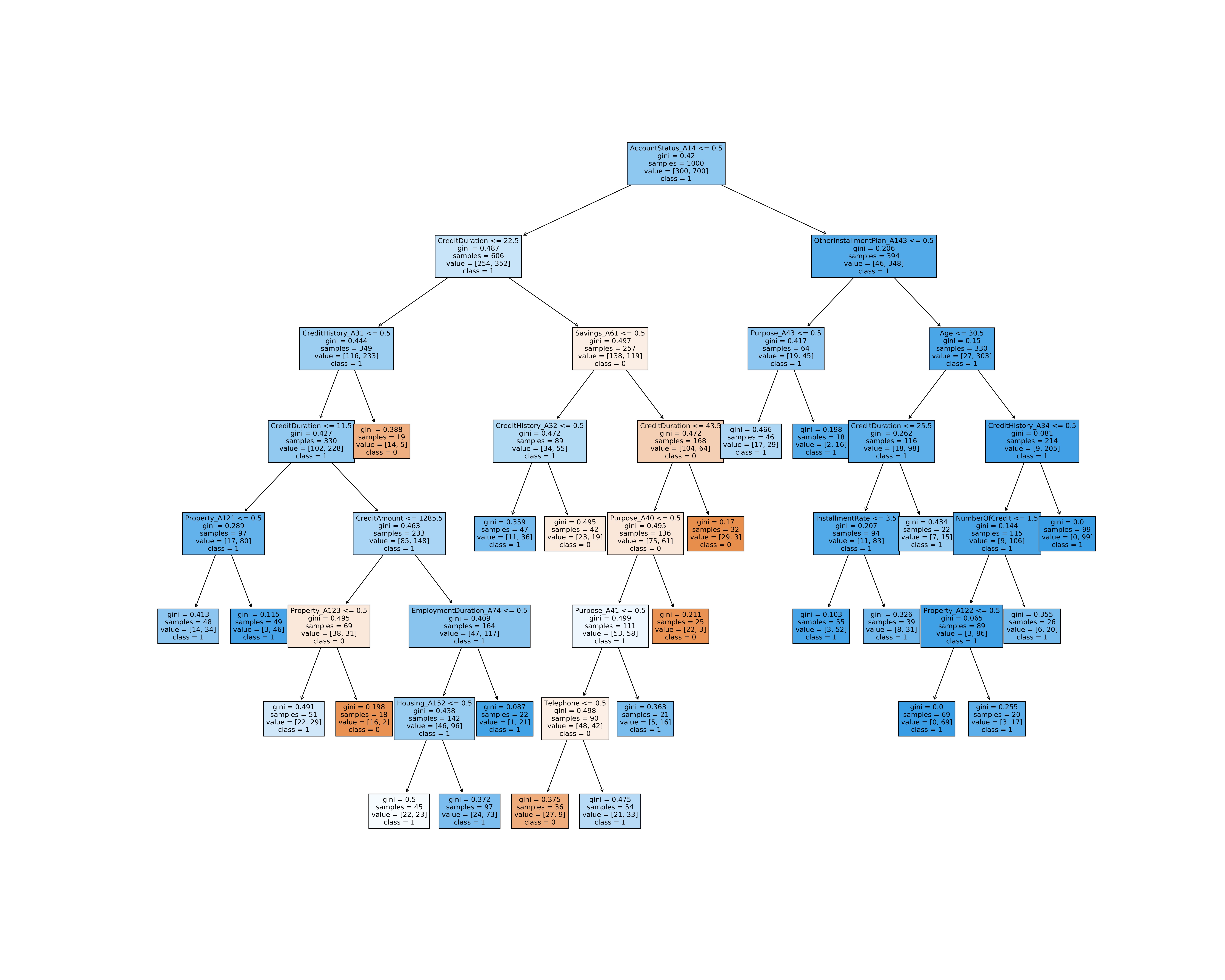} 
		\vskip 0.3cm 
		\label{Figure_Fairness_Decision} 
	\end{figure}
	\vspace{-0.5cm}
	{\noindent \footnotesize Notes: This figure displays the decision tree obtained with the hyperparameters mentionned in the TREE-prime column of Table \ref{Appendix_HyperparameterA}.}

    \clearpage
    
	\subsection{FPDP analysis for TREE-prime model}
	
	\label{Appendix_FPDP_TREE}
	
	\begin{figure}[!htbp] 
		\centering 
		\caption{Fairness PDP for conditional statistical parity in TREE-prime model} 
		\includegraphics[width=1.1\textwidth,trim={4.5cm 6cm 0 6cm},clip]{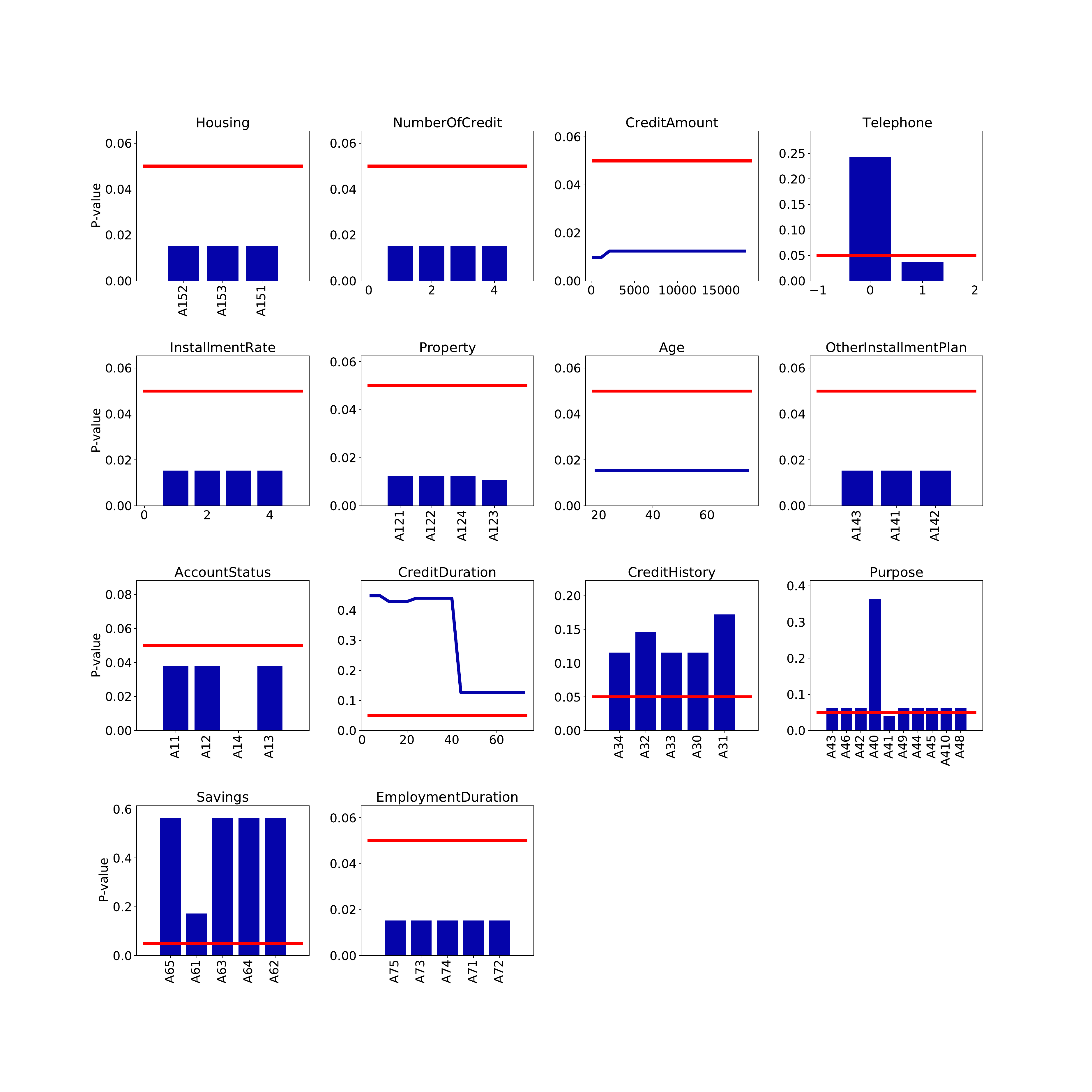} 
		\vskip 0.3cm 
		\label{fig:FPDP_CP} 
	\end{figure}\noindent {\footnotesize Notes: Each subplot displays the FPDP for conditional statistical parity, associated to a given feature and the classification TREE-prime model. The Y-axis displays the p-value of the conditional statistical parity statistic. The red line represents the 5\% threshold.}
	\newpage 
	\begin{figure}[!htbp] 
		\centering 
		\caption{Fairness PDP for equal odds in TREE-prime model} 
		\includegraphics[width=1.10\textwidth,trim={4.5cm 6cm 0 6cm},clip]{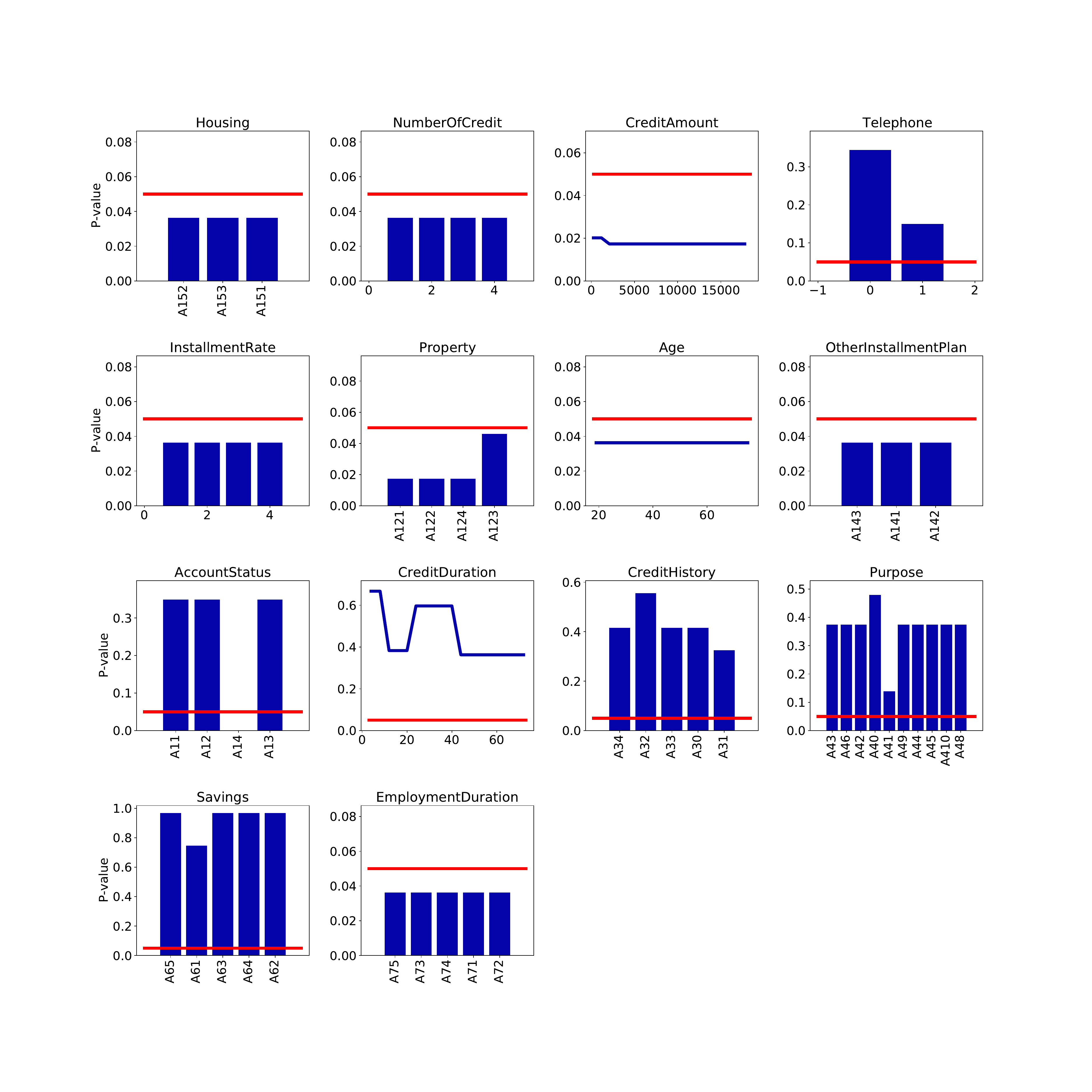} 
		\vskip 0.3cm 
		\label{fig:FPDP_EO} 
	\end{figure}\noindent {\footnotesize Notes: Each subplot displays the FPDP for equal odds, associated to a given feature and the classification TREE-prime model with indirect discrimination. The Y-axis displays the p-value of the equal odds test statistic. The red line represents the 5\% threshold.}
	\clearpage 
	\begin{figure}[!htbp] 
		\centering 
		\caption{Fairness PDP for equal opportunity in TREE-prime model} 
		\includegraphics[width=1.10\textwidth,trim={4.5cm 6cm 0 6cm},clip]{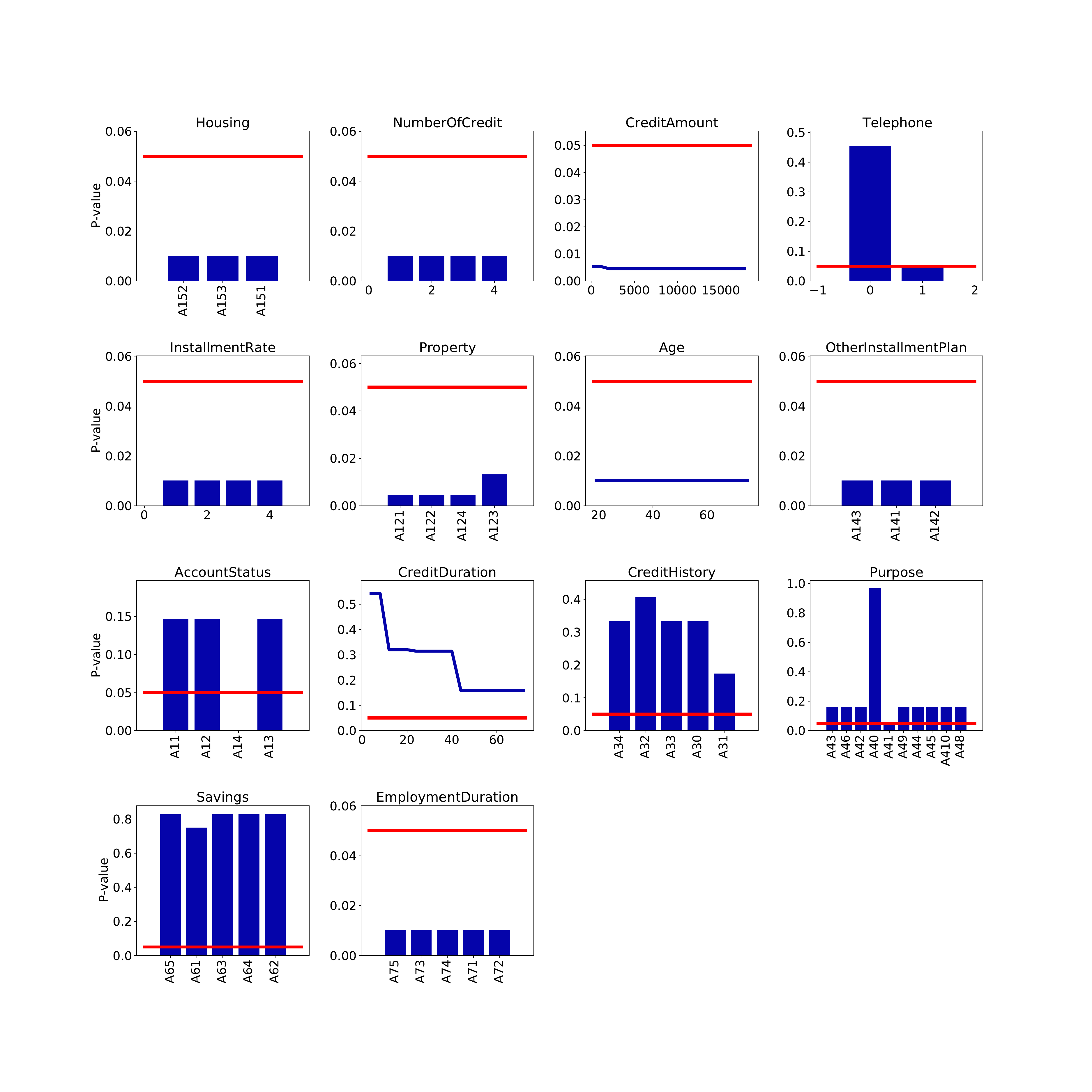} 
		\vskip 0.3cm 
		\label{fig:FPDP_EOP} 
	\end{figure}\noindent {\footnotesize Notes: Each subplot displays the FPDP for equal opportunity, associated to a given feature and the classification TREE-prime model. The Y-axis displays the p-value of the equal opportunity test statistic. The red line represents the 5\% threshold.}

    \clearpage

    \subsection{Analysis of the fairness cost for TREE-prime model}

        \begin{figure}[!h] 
    \centering 
    \caption{Misclassification cost-fairness trade-off} 
    \bigskip
    \includegraphics[width=0.9\textwidth,trim={0cm 0cm 0 0cm},clip]{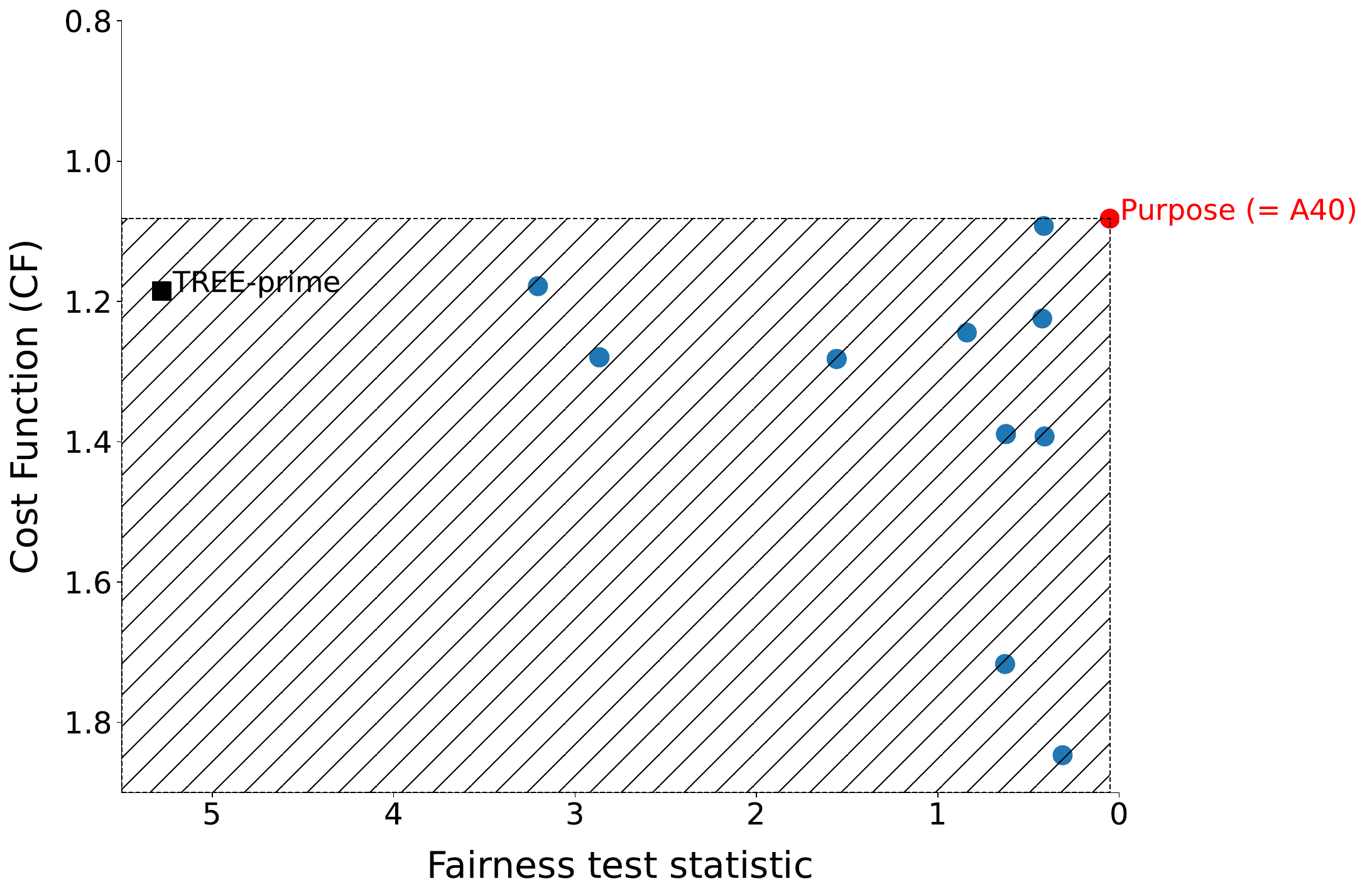} 
    \label{Figure Fairness vs Costs German} 

    \justifying
    \noindent {\footnotesize Notes: This figure displays the relationship between the misclassification cost $CF$ and the fairness test statistic associated with the null hypothesis of statistical parity, for the modified TREE-prime models and after mitigation. Each point represents a modified model linked to a candidate variable and a specific value. The values correspond to those reported in Panel B of Table \ref{tab:Mitigation}. Red dots correspond to Pareto optimal models that surpass the others, both in terms of costs and fairness. A model outperforms another one in terms of fairness if its test statistic is lower than the test of the other model.}
\end{figure}

\clearpage

\begin{figure}[!h] 
    \centering 
    \caption{Expected percentage increase in misclassification costs} 
    \bigskip
    \includegraphics[width=0.9\textwidth,trim={0cm 0cm 0 0cm},clip]{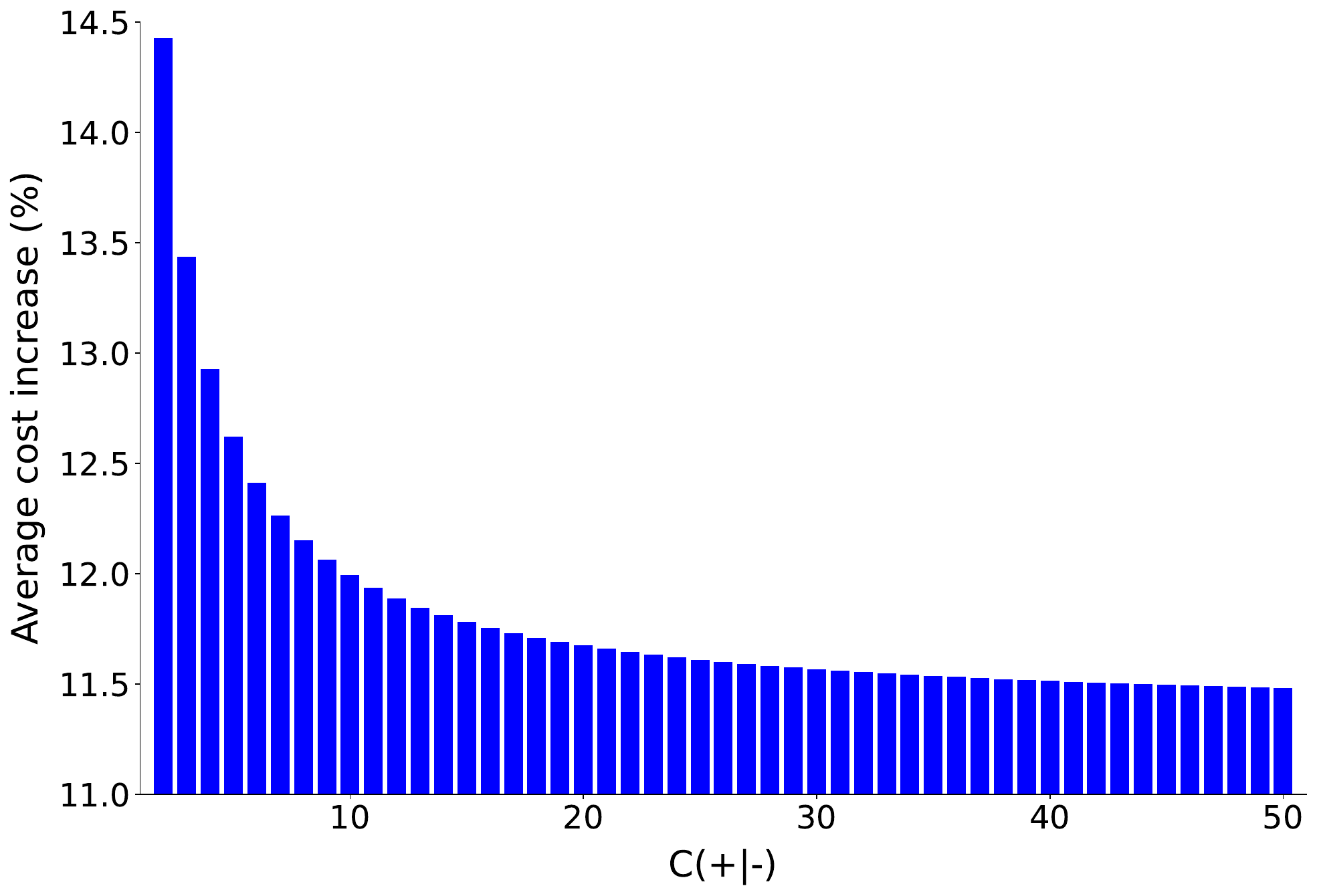} 
    \label{Figure cost ratios German}

    \justifying
    \noindent {\footnotesize Notes: This figure displays the expected percentage increase in error costs compared to the TREE-prime model across cost levels $C_{+|-}$ assuming $C_{-|+}=1$. Each line corresponds to a specific mitigation method consisting of "muting" the corresponding candidate variable.}
    \end{figure}

	\clearpage 

    \subsection{The effect of unfairness mitigation on loan approval}

    \label{section:loansgranted}

    \noindent Mitigating unfairness can have a significant impact on lenders' actions. When one removes a candidate feature to address a lack of fairness, it unavoidably modifies credit-granting decisions. For example, in a simple logistic regression model, muting a candidate variable by fixing its value is tantamount to altering the intercept or, similarly, as a change in the probability threshold splitting applicants between good and bad types. Changing this threshold also affects the unconditional probability to grant a loan, so the results of panels A and B of Table \ref{Table 1} are not fully comparable.\footnote{In our logistic regression model, the threshold value $\delta$ has been fixed such that the predicted proportion of defaults is equal to the actual one, i.e., $\Pr(\hat{Y}=0)=\Pr(Y=0)=0.3$.} However, as shown in the next paragraph, when we control for this probability by changing the threshold, our mitigation method remains valid in terms of fairness diagnosis and gives similar results in terms of accuracy. In the case of non-linear ML methods, the analysis is less obvious, but our experiments demonstrate that this adjustment does not impair the effectiveness of our mitigation method (see Table \ref{tab:Mitigation1}).\medskip

    \noindent Consider a logistic regression model for which the score, i.e., the probability to be classified as good type, is given by $P(X_i,C_i)=\Pr({Y}_{i}=1|X_i,C_i)=\Lambda(X_{i}\beta+\gamma C_{i}+\alpha)$, where $X_{i}$ is a set of features assumed to be independent of the protected attribute $D_{i}$, $C_{i}$ is a candidate variable correlated with $D_{i}$, and $\Lambda(.)$ is the cdf of a logistic distribution. The decision rule of the bank is given by $\widehat{Y}=1$ if $P(X_i,C_i)>\tau$ or equivalently if $X_{i}\beta+\gamma C_{i}+\alpha>\delta$ with $\delta=\Lambda^{-1}(\tau)$. For simplicity, we consider a binary candidate variable $C_{i}$ that takes value $1$ with probability $q$ and $-1$ with probability $1-q$. Finally, assume that the mitigation method consists in setting the same value $C_{i}=1$ for all credit applicants $i=1,\dots,n$. In this context, the "neutralisation" of the candidate variable can be viewed as a shift of the constant of the model from $\alpha$ to $\alpha+\gamma$, and then of the individual scores $P({X}_{i},{C}_{i})$. Therefore, it is true that our mitigation method changes the probability of granting a loan in the population. Alternatively, the bank could adjust the threshold such that the number of loans stays constant compared to the version of the model without neutralization. Formally, the new decision rule becomes $\widehat{Y}=1$ if $P({X}_{i},{C}_{i})>\tilde{\delta}$ with: 
    \begin{equation*}
        \tilde{\delta}=G_{X}^{-1}(q \cdot G_{X}(\alpha+\gamma-\delta)+(1-q) \cdot G_{X}(\alpha-\gamma-\delta))-\gamma-\alpha
    \end{equation*}
    where $G_{X}(.)$ is the cdf of the probability distribution of the index $X_{i}\beta$, which is assumed to be symmetric for simplicity. However, this change will not impact the efficiency of our mitigation method. For instance, if we consider the statistical parity definition, the probability difference between women ($D_{i}=1$ and $C_{i}=1$) and men ($D_{i}=0$ and $C_{i}=0$) before mitigation is:
    \begin{equation*}
        \Delta_{i}(\delta)=\Pr(\widehat{Y}_{i}=1|D_{i}=1)-\Pr(\widehat{Y}_{i}=1|D_{i}=0)=G_{X}(\alpha+\gamma-\delta)-G_{X}(\alpha-\delta) \ne 0
    \end{equation*}
    After mitigation with $C_{i}=1$, this difference is null while the probability to grant a loan in the population still remains unchanged through the use of the new threshold $\tilde{\delta}$:
    \begin{align*}                         
        \Delta_{i}(\tilde{\delta},C_{i}=1) &= \Pr(\widehat{Y}_{i}=1|D_{i}=1,C_{i}=1)-\Pr(\widehat{Y}_{i}=1|D_{i}=0,C_{i}=1) \\
        &= G_{X}(\alpha+\gamma-\tilde{\delta})-G_{X}(\alpha+\gamma-\tilde{\delta})=0                         
    \end{align*}
    In a more general framework with non linear ML models, changing the probability threshold to keep unchanged the numbers of loans granted will also modify the fairness metric and the AUC. However, we find that it does not affect the efficiency of our mitigation method. In Table \ref{tab:Mitigation1}, we report the fairness test statistic (statistical parity), percentage of correct classification (PCC), false discovery rate (FDR), misclassification costs (labeled CF, see below), and the number of approved loans for the TREE-prime model before and after mitigation, for several "muted" features. We consider two scenarios: case (1) where the probability threshold remains unchanged, and case (2) where the probability threshold is adjusted to keep the number of approved loans identical to the pre-mitigation situation.\footnote{Originally, there were 828 approved loans. Given the non-smooth nature of probabilities and the effective discrimination performance of our classification tree models (the predicted probabilities tend to be extremely low or high), it is not always feasible to achieve the exact same number of approved loans after adjusting the probability threshold. We identify the probability threshold that minimizes the absolute difference between the number of approved loans before and after mitigation.} In essence, the results show that it is always possible to find a combination of mitigation feature and probability threshold that ensures compliance with fairness tests while maintaining the number of loans and accuracy measures approximately constant. For instance, if the bank prioritizes the PCC, setting the Purpose variable to modality A49 (business) enables achieving a PCC of $76.8\%$ (vs. $79\%$ before mitigation) while maintaining the number of approved loans virtually constant ($830$ vs. $828$). We have shown that our key findings remain valid. Indeed, it is possible to completely eradicate discrimination bias while incurring only minor accuracy trade-offs and maintaining the current lending activity.

    \begin{sidewaystable}[h] 
     \centering 
     \caption{Mitigation} 
     \label{tab:Mitigation1} 
     \vspace{2mm} 
    
    \subcaption*{Panel A: With re-estimation}
    \begin{tabular}{p{4.1cm}llccccccccc}
    \toprule
    {} &  \multicolumn{1}{c}{SP}  & \multicolumn{1}{c}{SP} &   \multirow{2}{*}{AUC} &   PCC & PCC &  FDR & FDR &  CF & CF & Loans & Loans \\
    {} &  \multicolumn{1}{c}{(1)}  &  \multicolumn{1}{c}{(2)} &   &   (1) & (2) & (1)  & (2) & (1) & (2) & (1) & (2) \\
    \midrule
    TREE-prime      &              0.0216*  & \multicolumn{1}{c}{X} & 0.8393 &  79.0 & X &  0.2041 & X & 1.1852 & X &  828 & X \\
    \midrule
    Telephone      &              0.1019 & 0.1019  & 0.8380 &  78.7 & 78.7 &  0.2041 & 0.2041 & 1.1843 & 1.1843 & 823 & 823 \\
    CreditHistory  &              0.0462* & 0.0462* & 0.8336 &  78.2 & 78.2 &  0.2061 & 0.2061 & 1.1967 & 1.1967 & 820 & 820 \\
    CreditDuration &              0.6463 & 0.4467  & 0.8017 &  75.5 & 75.5 &  0.2153 & 0.2288 &  1.2510 & 1.3557 & 799 & 839 \\
    Purpose        &              0.0263* & 0.0608 & 0.7804 &  74.6 & 74.6 &  0.2170 & 0.2274 &  1.2586 & 1.3371 & 788 & 818 \\
    Savings        &              0.0332* & 0.0559 & 0.7130 &  72.7 & 72.1 &  0.2615 & 0.2332 & 1.6157 & 1.3624 & 895 & 789 \\
    AccountStatus  &              0.9875 & 0.1281  & 0.6727 &  71.8 & 70.4 &  0.2743 & 0.2518 & 1.7333 & 1.4967 & 926 & 814 \\
    \bottomrule
    \end{tabular}
    \medskip
    \subcaption*{Panel B: Without re-estimation}
    \begin{tabular}{lllccccccccc}
    \toprule
    \multirow{2}{*}{} & \multicolumn{1}{c}{SP} & \multicolumn{1}{c}{SP} & \multirow{2}{*}{AUC} & PCC & PCC & FDR & FDR & CF & CF & Loans & Loans \\  
    & \multicolumn{1}{c}{(1)} & \multicolumn{1}{c}{(2)} &  & (1) &  (2) &  (1) &  (2) &  (1) &  (2) & (1) &  (2) \\ 
    \midrule
    TREE-prime      &              0.0216* & \multicolumn{1}{c}{X} & 0.8393 &  79.0  & X &    0.2041  & X &  1.1852 & X & 828 & X \\
    \midrule
    Telephone (= 0) & 0.5195 & 0.5195 & 0.8325 & 77.8 & 77.8 & 0.1912 & 0.1912 & 1.0924 & 1.0924 & 774 & 774 \\
    Purpose (= A49) & 0.0905 & 0.0905 & 0.8219 & 76.8 & 76.8 & 0.2181 & 0.2181 & 1.2795 & 1.2795 & 830 & 830 \\
    Savings (= A61) & 0.5150 & 0.6339 & 0.8212 & 77.0 & 76.1 & 0.2106 & 0.2226 & 1.2243 & 1.3105 & 812 & 831\\
    Purpose (= A40) & 0.8206 & 0.1783 & 0.8191 & 76.7 & 75.4 & 0.1899 & 0.2211 & 1.0819 & 1.2943 & 753 & 814 \\
    Purpose (= A43) & 0.0905 & 0.0905 & 0.8171 & 76.8 & 76.8 & 0.2181 & 0.2181 & 1.2795 & 1.2795 & 830 & 830 \\
    CreditHistory (= A32) & 0.3596 & 0.3596 & 0.8099 & 75.6 & 75.6 & 0.2143 & 0.2143 & 1.2443 & 1.2443 & 798 & 798 \\
    CreditHistory (= A34) & 0.5212 & 0.0207* & 0.8068 & 77.7 & 77.6 & 0.2311 & 0.2167 & 1.3924 & 1.2733 & 887 & 840 \\
    Savings (= A65) & 0.4296 & 0.4296 & 0.7753 & 73.9 & 73.9 & 0.2346 & 0.2346 & 1.3890 & 1.3890 & 827 & 827 \\
    AccountStatus (= A13) & 0.0734 & 0.2326 & 0.7418 & 72.9 & 75.2 & 0.2066 & 0.2310 & 1.1781 & 1.3705 & 731 & 840 \\
    CreditDuration (= 20.0) & 0.4277 & 0.4277 & 0.7408 & 72.6 & 72.6 & 0.2715 & 0.2715 & 1.7167 & 1.7167 & 932 & 932 \\
    CreditDuration (= 24.0) & 0.2120 & 0.7265 & 0.7377 & 68.2 & 68.2 & 0.2264 & 0.2264 & 1.2819 & 1.2819 & 698 & 698 \\
    CreditDuration (= 8.0) & 0.5767 & 0.5767 & 0.7197 & 71.2 & 71.2 & 0.2854 & 0.2854 & 1.8467 & 1.8467 & 960 & 960 \\
    \bottomrule
    \end{tabular}

    \label{tab:Mitigation_all}
		
  \medskip \justifying
	\noindent {\footnotesize Notes: This table reports the mitigation results obtained with re-estimation (Panel A) and without re-estimation (Panel B) of the TREE-prime model. Panel A displays the p-values of the fairness test according to Statistical Parity (SP) and the performance metrics (AUC, PCC, FDR, CF) obtained after removing one candidate variable from the list of explanatory variables. In this case, the model is re-estimated. Panel B displays the p-values of the fairness test according to Statistical Parity (SP) and the performance metrics (AUC, PCC, FDR, CF) obtained after setting one candidate variable to the same value for all individuals. The value of the candidate variable is reported in parentheses. In this case, the model is not re-estimated. We differentiate between settings in which (1) we do not modify and those in which (2) we do modify the model's threshold. We modify the threshold to ensure that the number of loans granted by the bank remains constant before and after the mitigation process. The last two columns display the number of loans granted by the bank in the two settings.}
 \end{sidewaystable}

    \clearpage

    \subsection{Honest estimation procedure}

    \label{appendix:honest_estimation}
    Using the same dataset for both training the algorithm and assessing its fairness may alter the external validity of the fairness diagnosis. By external validity, we refer here to the fact that the fairness diagnosis for a given scoring model should remain unchanged when applied to a new sample. The fairness diagnosis may vary if the distribution of the model's output $\hat{y}$ on a new dataset differs from that obtained on the training sample. This situation could arise if the model excessively learns from the training data, resulting in overfitting. This question is indeed becoming important in the literature on fairness \citep{Mandal2020}.

    \medskip
    \noindent To address this concern, we apply the "honest" estimation procedure introduced by \cite{Athey_2016}, where a first sample is used to construct the decision tree, a second one is employed to estimate the probability of default, and a third one is used to assess the fairness of the model outcomes. Using two separate samples to construct the decision tree and to estimate the probability of default in each leaf allows us to limit the overfitting of the model and as a consequence, to increase the external validity of our fairness test results.
    
    \medskip
    \noindent We now define the notations associated to this new estimation approach as in \cite{Athey_2016}. A tree $\Pi$ corresponds to a partitioning of the feature space $\mathcal{X}$, with $k=\#(\Pi)$ the number of elements in the partition. We write:
    \begin{equation*}
        \Pi = \{\ell_1,...,\ell_k\}, \mbox{ with } \cup_{j=1}^{k}\ell_j = \mathcal{X}
    \end{equation*}
    where $\ell_j=\ell_j(x;\Pi)\in \Pi$ denotes a leaf and $x \in \ell_j$ the features of a loan applicant in leaf $\ell_j$. Let $\mathbb{S}$ be the space of samples from a population and $\mathbb{P}$ the space of partitions. Let $\pi: \mathbb{S} \rightarrow \mathbb{P}$ be an algorithm that, on the basis of a sample $\mathcal{S} \in \mathbb{S}$, constructs a decision tree. 
    
    \medskip
    \noindent The so-called "honest" estimation procedure relies on a training sample $\mathcal{S}^{tr}$, an estimation sample $\mathcal{S}^{est}$, and a test sample $\mathcal{S}^{test}$, and it proceeds in three steps. First, it uses the training sample $\mathcal{S}^{tr}$ to build the decision tree and the corresponding partition $\Pi^{tr}$.\footnote{Contrary to \cite{Athey_2016}, we refrain from altering the splitting and cross-validation criteria, specifically the Gini criterion in our case. Two reasons underpin this decision. Firstly, we regard the credit scoring model as pre-trained and given, and modifying the tree structure after it has been validated by internal control teams and supervisors is not realistic. Secondly, in contrast to \cite{Athey_2016}, who adjust the splitting criteria for treatment effect estimation, we do not use the fairness metric as the splitting criterion. Indeed, our objective is not to impose a fairness constraint during the estimation of the model but to assess the fairness of the model post-estimation.} Second, it uses the estimation sample $\mathcal{S}^{est}$ to compute the conditional probability for an applicant of being good type in each leaf of the tree. Note that in a standard approach, the training and estimation samples are identical. Finally, the test sample $\mathcal{S}^{test}$ is used to assess the fairness of the model by evaluating the outcomes $\hat{Y}$. We denote $D^{test}$ the protected attribute observed on the test sample and $\hat{Y}^{test}(X^{test},\mathcal{S}^{est},\Pi^{tr})$ the predicted outcome obtained from the features $X^{test}$, the partition $\Pi^{tr}$, and the probabilities estimated with the dataset $S^{est}=(X^{est},Y^{est})$. Then, the fairness test statistic is defined as follows: 
        \begin{equation*}
            F_{H_{0,i}}(\mathcal{S}^{test},\mathcal{S}^{est},\mathcal{S}^{tr}) \equiv h_{i}\left(\widehat{Y}_{j}(X_j,\mathcal{S}^{est},\Pi^{tr}),Y_{j}^{test},D_{j}^{test};j \in \mathcal{S}^{test}\right)
        \end{equation*}
    where $h_{i}$ denotes a functional form that depends on the null hypothesis $H_{0,i}$. One key advantage of the honest estimation method is that the three steps of the procedure closely resemble the three phases of the PD modeling typically used by banks \citep{EBA2017}, namely risk differentiation, risk quantification, and model backtesting.\footnote{Risk differentiation involves using a model to estimate a continuous score based on risk factors and then categorizing borrowers or credit exposures into different homogeneous risk classes based on this score. The risk quantification step involves assigning a default probability to each homogeneous risk class. Generally, these probabilities are estimated by the empirical frequency of defaults observed over a long period within each class. The risk differentiation and quantification phases typically involve two different datasets, similar to the honest estimation procedure. Finally, the backtesting of the model involves using a test sample different from the two previous ones.}
    
    \medskip
    \noindent We implement the honest estimation procedure on the Taiwan credit database including $25,983$ individuals. This enables us to clearly distinguish between the samples used for model training, estimation, and fairness assessments.\footnote{We do not apply the honest estimation to the German credit database as it only contains $690$ male applicants and $310$ female applicants. As a result, it poses a challenge to divide the dataset into multiple sets without facing the risk of having too few individuals for certain tests, such as conditional statistical parity tests.} We then compare the results of our fairness tests using or not the honest estimation. This analysis produces two key findings. Firstly, we corroborate your assertion that utilizing the same dataset for both training the algorithm and assessing its fairness may alter the external validity of the fairness diagnosis. Secondly, we notice that employing the standard train/test split procedure yields identical fairness diagnosis and comparable predictive performances (AUC, PCC, FDR) compared to the honest estimation procedure. This finding implies that implementing a usual train-test splitting approach enables to get dependable fairness diagnosis, signifying that the diagnosis remains consistent when applied to a new subsample. 
    
    \medskip
    \noindent In detail, we first divide the dataset (25,983 observations) into three distinct samples: a training sample of 17,408 observations (67\%), an estimation sample of 6,500 observations (25\%), and a test sample of 2,075 observations (8\%). Next, we assess the fairness of a decision tree using three different procedures: TREE-all, TREE, and TREE-honest. In all cases, we calculate the fairness tests and the accuracy measures (AUC, PCC, FDR) on the test sample. However, in the first case (TREE-all), the decision tree is constructed using the combined training/estimation/test sample. In this setup, we evaluate fairness on the data employed for training the model. In the second case (TREE), the model is built using both the training and estimation samples, excluding the test sample. For the last case (TREE-honest), the partition is generated from the training sample, while the outcome probabilities are calibrated using the estimation sample. The second configuration aligns with the conventional train/test split procedure, while the last one corresponds to the honest estimation procedure. 
    
    \medskip
    \noindent The results are presented in Table \ref{honest_estimation}. At a 10\% significance level, we reject the null hypothesis of statistical parity and conditional parity for the TREE-all model (column 1), while we do not reject the null hypothesis for the TREE (column 2) and TREE-honest (column 3) models. This discrepancy serves as a compelling example of your intuition regarding the lack of external validity of the fairness diagnosis obtained from a sample used to train the algorithm. We also observe that the fairness diagnosis and accuracy measures of the TREE and TREE-honest models are similar. indeed, for both models, we do not reject the null hypothesis of fairness. Moreover, the PCC, FDR, and CF are very close across the two models. For instance, the PCC of the TREE model is equal to 79.81\% and 78.84\% for the TREE-honest. However, we note that the AUC of the TREE-honest is lower than the one for TREE (0.6926 vs. 0.7296). 
    
    \medskip
    \noindent These results suggest that employing a standard train/test split procedure appears to be sufficient to yield dependable fairness test diagnoses and to control overfitting. 
    
    \clearpage
    
    \begin{table}[h] 
     \centering 
     \caption{Fairness tests with and without honest estimation} 
     \label{honest_estimation} 
    \begin{tabular}{l c c c}
     \toprule 
    {} & \multicolumn{1}{c}{TREE-all} & \multicolumn{1}{c}{TREE} & \multicolumn{1}{c}{TREE-honest} \\
    \bottomrule 
    Statistical parity & 0.0688* & 0.6393 &  0.7778  \\
    Cond. stat. parity & 0.0587*	 & 0.6698 &  0.8153 \\
    Equal odds & 0.2912 & 0.5183 &  0.7926 \\
    Equal opportunity & 0.1722 & 0.3736 &  0.6918 \\
    Predictive equality & 0.4372 & 0.4698 &  0.5791 \\
    Calibration & 0.3484 & 0.1752 &  0.4507 \\
    \midrule 
    AUC & 0.7906 & 0.7296 & 0.6926 \\
    PCC & 80.19 & 79.81 & 78.84 \\
    FDR  & 0.1774 & 0.1793 & 0.1795 \\
    CF & 1.3880 & 1.4036 & 1.3987  \\
    \bottomrule 
    \end{tabular}
    
    \medskip \justifying \noindent \footnotesize {Notes: This table displays the accuracy measures and p-values of the fairness tests obtained for three different decision trees. We use three distinct datasets: a training sample, an estimation sample, and a test sample. The first decision tree (TREE-all) is constructed using all three samples combined. The second decision tree (TREE) is built using the training and estimation samples. The third one (TREE-honest) employs the training sample to generate the partition, and subsequently estimates the probability of default using the estimation sample. The results presented in this table are derived from the test sample. ** indicates statistical significance at 5\%, * at 10\%.}
    \end{table}

    \clearpage

    \subsection{External validity: Taiwan credit database}

	\begin{table}[h] 
     \centering 
     \caption{Model performances without the protected feature} 
     \label{tab:perf_taiwan} 
     \vspace{2mm} 
    \begin{tabular}{c c c c c c c}
    \toprule 
    {} & LR & TREE & RF & XGB & SVM & ANN \\
    \midrule 
    AUC & 0.7335 & 0.7488 & 0.7711 & 0.7701 & 0.7430 & 0.7692 \\
    PCC & 78.31 & 80.58 & 81.38 & 81.22 & 81.26 & 81.07 \\
    FDR & 0.1427 & 0.1724 & 0.1726 & 0.1707 & 0.1731 & 0.1683 \\
    CF & 1.0768 & 1.3426 & 1.3491 & 1.3300 & 1.3532 & 1.3058 \\
    \bottomrule \\ 
    \end{tabular} 
    
    \medskip \justifying \noindent {\footnotesize Notes: This table reports the area under the ROC curve (AUC), the percentage of correct classification (PCC), the False Discovery Rate (FDR), and a Cost Function (CF) values for each scoring model, without gender. The CF is defined as a weighted average of type-I and type-II errors, where the weight associated to the type-I (type-II) error is equal to 2 (1). LR: Logistic Regression, TREE: classification tree, RF: Random Forest, XGB: XGBoost, SVM: Support Vector Machine, ANN: Artificial Neural Network.}
     \end{table}

    \bigskip 
    
    \begin{table}[h!]
    \centering
    \caption{Fairness tests for models without gender} 
     \vspace{2mm}\label{Table 3 Taiwan} 
    \begin{tabular}{lllllll}\toprule 
    {} &       \multicolumn{1}{c}{LR} &     \multicolumn{1}{c}{TREE} &      \multicolumn{1}{c}{RF}  &   \multicolumn{1}{c}{XGB}    &  \multicolumn{1}{c}{SVM}     &  \multicolumn{1}{c}{ANN}     \\
    \midrule
    Statistical parity & 0.1440 & 0.1114 & 0.0691 & 0.0218* & 0.1971 & 0.0494* \\
    Cond. parity & 0.2056 & 0.3129 & 0.2138 & 0.0822 & 0.4729 & 0.1486 \\
    Equal odds & 0.9837 & 0.2201 & 0.1681 & 0.1254 & 0.1904 & 0.3993 \\
    Equal opportunity & 0.9923 & 0.1346 & 0.0821 & 0.0445* & 0.1662 & 0.1851 \\
    Predictive equality & 0.8563 & 0.3743 & 0.4610 & 0.7344 & 0.2368 & 0.7775 \\
    Calibration & 0.0815 & 0.0613 & 0.2102 & 0.1538 & 0.0614 & 0.1461 \\\bottomrule 
    \end{tabular}
    
    \medskip \justifying
    \noindent {\footnotesize Notes: This table reports the p-values of the fairness tests (see Section \ref{Section_Fairness_Inference}) obtained for the scoring models without using the gender variable. * indicates statistical significance at 5\%. LR: Logistic Regression, TREE: classification tree, RF: Random Forest, XGB: XGBoost, SVM: Support Vector Machine, ANN: Artificial Neural Network.}
    \end{table}

    \clearpage
    
    \begin{figure}[!htbp] 
     \centering 
    \caption{Fairness PDP for the statistical parity in XGBoost model} 
    \includegraphics[width=1\textwidth,trim={4.5cm 6cm 0 6cm},clip]{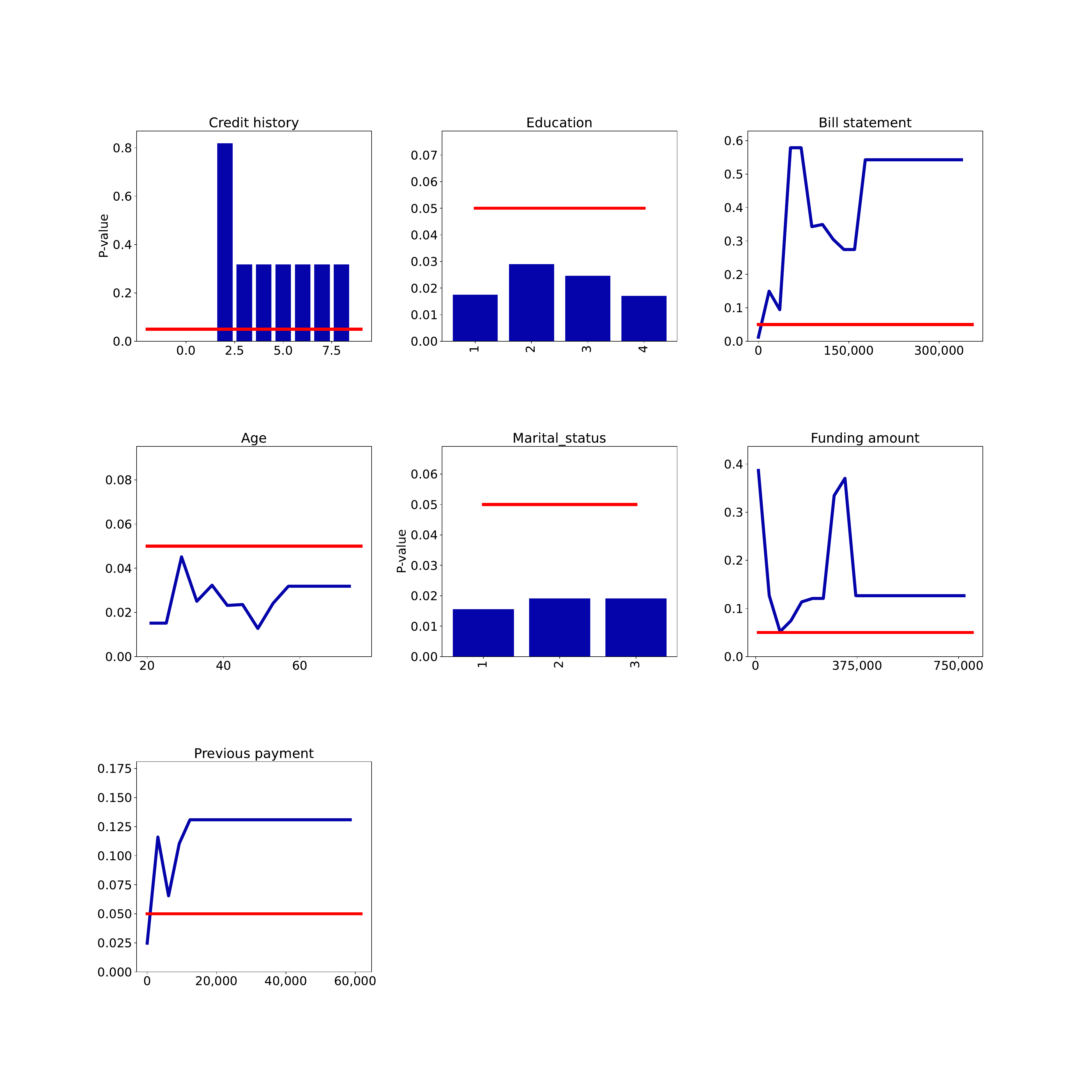} 
    \vskip 0.3cm 
    \label{FPDP Taiwan} 
    \medskip \justifying
    \noindent {\footnotesize Notes: Each subplot displays the FPDP for statistical parity, associated to a given feature and the classification XGBoost model. The Y-axis displays the p-value of the statistical parity test statistic. The red line represents the 5\% threshold.}
    \end{figure}

    \clearpage

    \begin{figure}[!htbp] 
     \centering 
    \caption{Measures of association between features, target variables, and gender} 
    \includegraphics[width=1\textwidth,trim={0cm 1cm 0 2cm},clip]{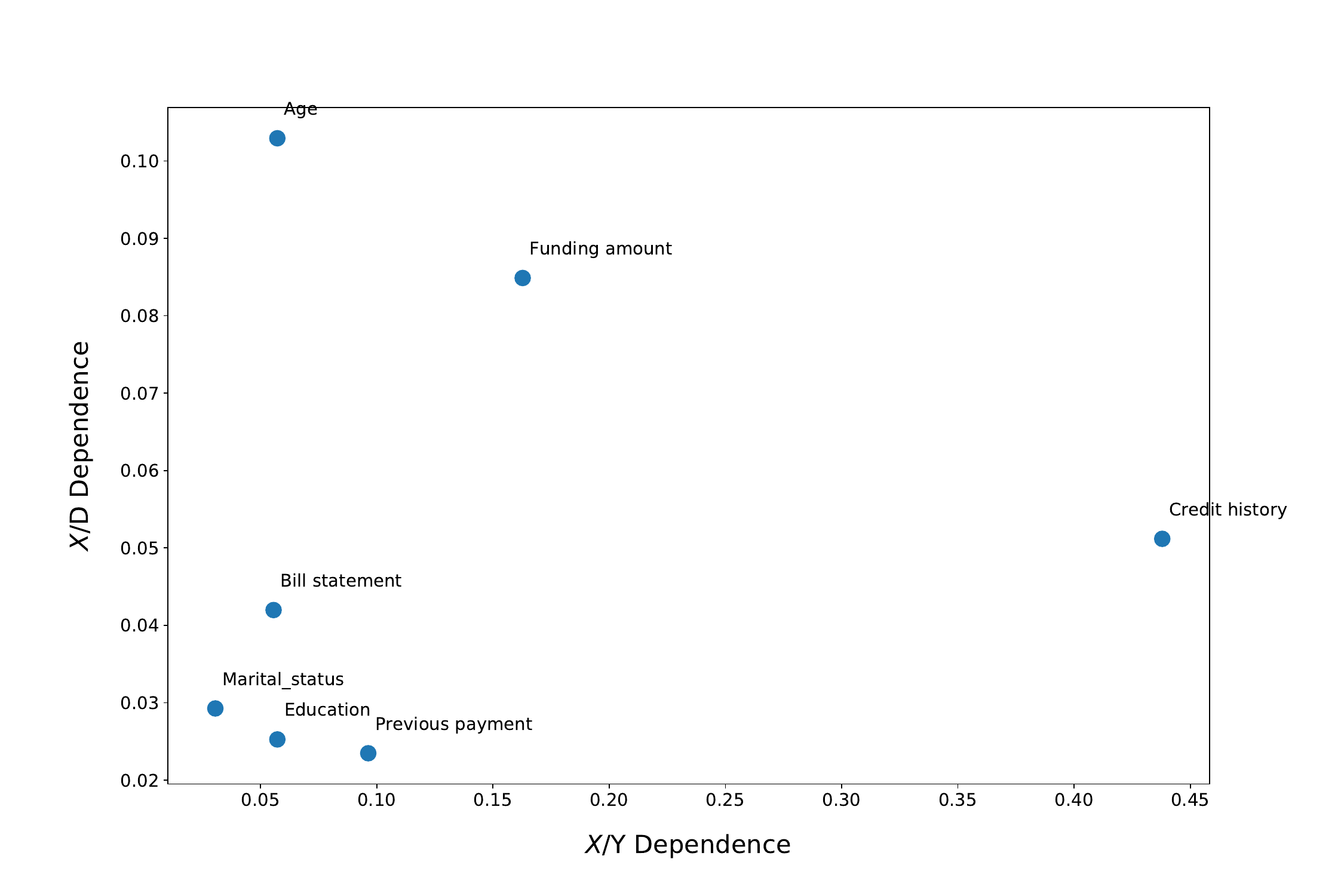}
    \label{VCramerTaiwan} 
    
    \medskip \justifying
    \noindent {\footnotesize Notes: This figure displays the Cramer's V measures between each feature and the default variable (horizontal axis) and the gender variable (vertical axis)}.
    \end{figure}

    \clearpage 

     \begin{table}[h] 
     \centering 
     \caption{Mitigation} 
     \label{tab:Mitigation_Taiwan} 
     \vspace{2mm} 
    
    \begin{tabular}{p{4.9cm}lcccc}
    \toprule
    \addlinespace[10pt]
    \multicolumn{6}{c}{Panel A: With re-estimation} \vspace{7pt} \\ \midrule
    {} &  \multicolumn{1}{c}{SP} &     AUC &    PCC &  FDR &    CF \\
    \midrule
    XGB              &              0.0218* &  0.7701 &  81.2 &    0.1707 &   1.3300 \\
    \midrule
    Bill statement & 0.1817 & 0.7602 & 81.26 & 0.1713 & 1.3363 \\
    Funding amount & 0.1066 & 0.7550 & 81.31 & 0.1715 & 1.3382 \\
    Previous payment & 0.1412 & 0.7398 & 81.31 & 0.1720 & 1.3433 \\
    Credit history & 0.2816 & 0.6579 & 76.93 & 0.2276 & 1.9204 \\
    \toprule
    \addlinespace[10pt]
    \multicolumn{6}{c}{Panel B: Without re-estimation} \vspace{7pt}\\ \midrule 
    {} &  \multicolumn{1}{c}{SP} &     AUC &    PCC &  FDR &    CF \\
    \midrule
    XGB              &              0.0218* &  0.7701 &  81.2 &    0.1707 &   1.3300 \\
    \midrule
    Previous payment (= 3,082) & 0.1161 & 0.7674 & 81.3 & 0.1724 & 1.3470 \\
    Previous payment (= 12,328) & 0.1309 & 0.7658 & 81.3 & 0.1738 & 1.3605 \\
    Previous payment (= 6,164) & 0.0654 & 0.7639 & 81.2 & 0.1740 & 1.3621 \\
    Previous payment (= 9,246) & 0.1104 & 0.7639 & 81.3 & 0.1740 & 1.3623 \\
    Bill statement (= 35,464) & 0.0943 & 0.7634 & 81.3 & 0.1719 & 1.3418 \\
    Bill statement (= 17,732) & 0.1498 & 0.7614 & 81.3 & 0.1719 & 1.3419 \\
    Bill statement (= 53,196) & 0.5786 & 0.7608 & 81.2 & 0.1754 & 1.3754 \\
    Bill statement (= 106,392) & 0.3493 & 0.7595 & 81.3 & 0.1733 & 1.3560 \\
    Bill statement (= 88,660) & 0.3426 & 0.7593 & 81.3 & 0.1734 & 1.3563 \\
    Bill statement (= 124,124) & 0.3046 & 0.7584 & 81.3 & 0.1727 & 1.3500 \\
    Bill statement (= 141,856) & 0.2743 & 0.7556 & 81.3 & 0.1726 & 1.3485 \\
    Bill statement (= 177,320) & 0.5428 & 0.7551 & 81.1 & 0.1711 & 1.3327 \\
    Funding amount (= 530,000) & 0.1265 & 0.7503 & 81.2 & 0.1752 & 1.3739 \\
    Bill statement (= 248,248) & 0.5428 & 0.7467 & 81.1 & 0.1711 & 1.3327 \\
    Funding amount (= 290,000) & 0.3344 & 0.7412 & 81.1 & 0.1772 & 1.3932 \\
    Funding amount (= 330,000) & 0.3703 & 0.7408 & 81.1 & 0.1775 & 1.3958 \\
    Funding amount (= 90,000) & 0.0519 & 0.7406 & 81.2 & 0.1734 & 1.3569 \\
    Funding amount (= 250,000) & 0.1207 & 0.7398 & 81.2 & 0.1738 & 1.3606 \\
    Funding amount (= 10,000) & 0.3867 & 0.7397 & 80.3 & 0.1653 & 1.2743 \\
    Funding amount (= 50,000) & 0.1269 & 0.7368 & 81.2 & 0.1744 & 1.3658 \\
    Funding amount (= 210,000) & 0.1207 & 0.7363 & 81.3 & 0.1738 & 1.3606 \\
    Funding amount (= 170,000) & 0.1136 & 0.7362 & 81.2 & 0.1750 & 1.3721 \\
    Funding amount (= 130,000) & 0.0742 & 0.7336 & 81.3 & 0.1736 & 1.3589 \\
    Credit history (= 3) & 0.3176 & 0.5360 & 26.6 & 0.1885 & - \\
    Credit history (= 2) & 0.8193 & 0.5203 & 27.3 & 0.2280 & - \\
    \bottomrule
    \end{tabular}
    \medskip

\label{tab:MitigationTaiwan}
\medskip \justifying
	\noindent {\footnotesize Notes: This table reports the mitigation results obtained with re-estimation (Panel A) and without re-estimation (Panel B) of the XGBoost model. Panel A displays the p-values of the fairness test according to Statistical Parity (SP) and the performance metrics (AUC, PCC, FDR, CF) obtained after removing one candidate variable from the list of explanatory variables. In this case, the model is re-estimated. Panel B displays the p-values of the fairness test according to Statistical Parity (SP) and the performance metrics (AUC, PCC, FDR, CF) obtained after setting one candidate variable to the same value for all individuals. The value of the candidate variable is reported in parentheses.}
 \end{table}

  \clearpage
    
    \begin{figure}[!htbp] 
     \centering 
    \caption{Accuracy-fairness trade-off} 
    \bigskip
    \includegraphics[width=0.9\textwidth,trim={0cm 0cm 0 0cm},clip]{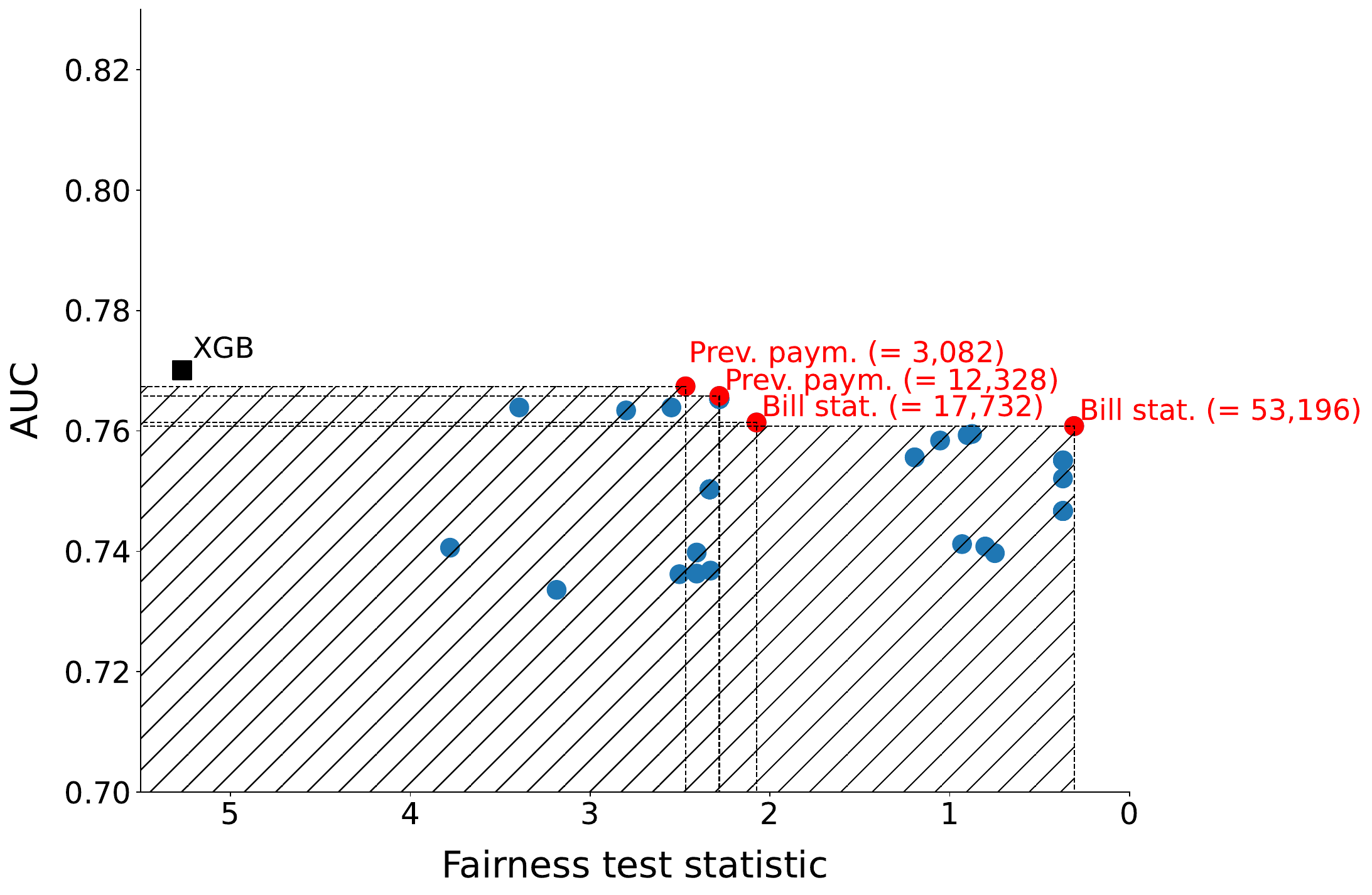} 
    
    \label{Figure 4 Taiwan} 
    \justifying
        \noindent {\footnotesize Notes: This figure displays the relationship between the AUC and the fairness test statistic associated with the null hypothesis of statistical parity, for the modified XGBoost models and after mitigation on the Taiwan dataset. Each point represents a modified model linked to a candidate variable and a specific value. The values correspond to those reported in Panel B of Table \ref{tab:MitigationTaiwan}. The red dots correspond to Pareto optimal models that dominate the others, both in terms of performance and fairness. A model dominates another one in terms of fairness if its test statistic is lower than the test of the other model.}
    \end{figure}

\clearpage

 \begin{figure}[!htbp] 
 \centering 
\caption{Misclassification cost-fairness trade-off} 
\bigskip
\includegraphics[width=0.9\textwidth,trim={0cm 0cm 0 0cm},clip]{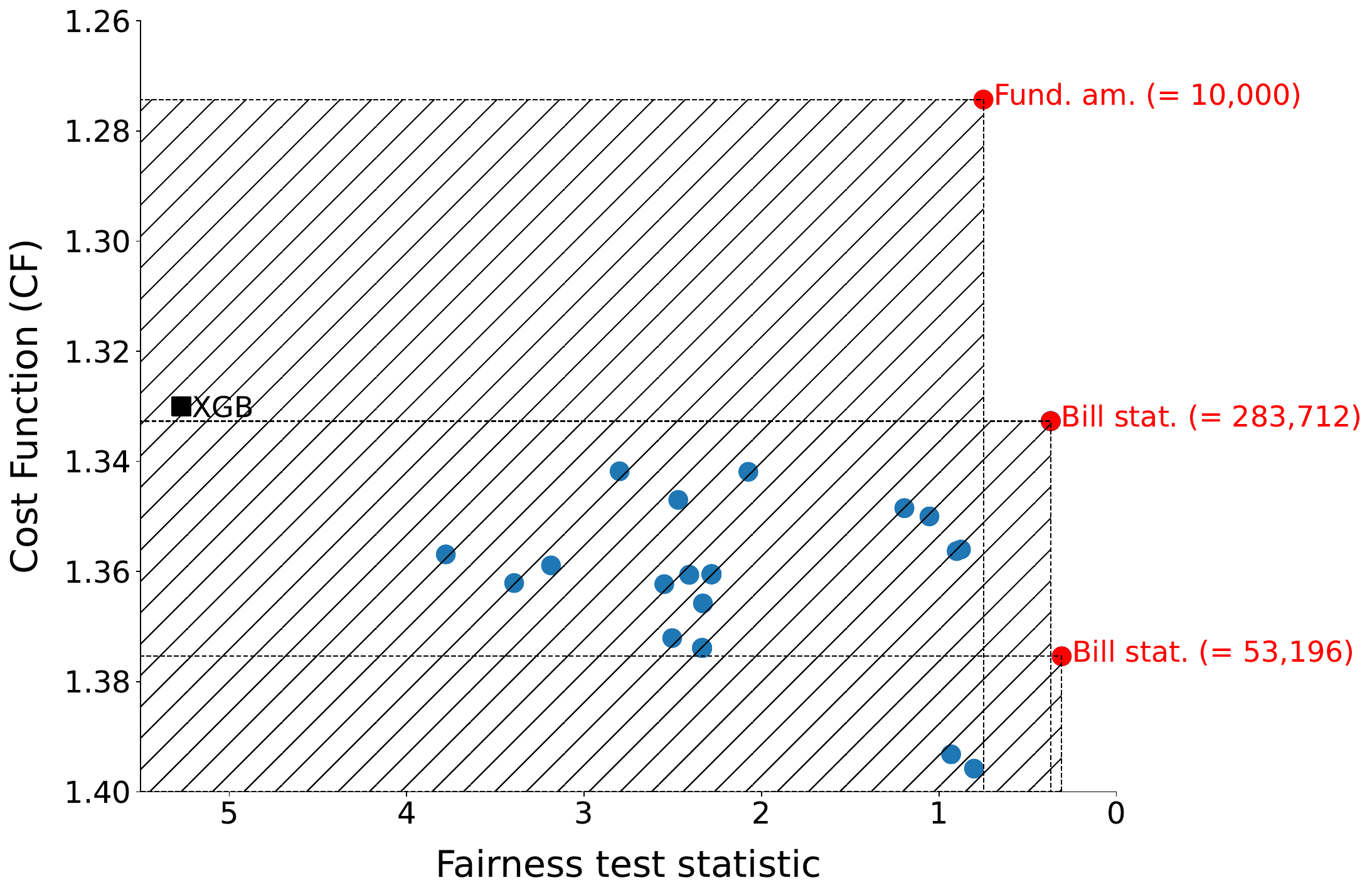} 

\label{Figure Fairness vs Costs Taiwan} 
\medskip \justifying
    \noindent {\footnotesize Notes: This figure displays the relationship between the misclassification cost $CF$ and the fairness test statistic associated with the null hypothesis of statistical parity, for the modified XGBoost models and after mitigation on the Taiwan dataset. Each point represents a modified model linked to a candidate variable and a specific value. The values correspond to those reported in Panel B of Table \ref{tab:Mitigation_Taiwan}. The red dots correspond to Pareto optimal models that dominate the others, both in terms of costs and fairness. A model dominates another one in terms of fairness if its test statistic is lower than the test of the other model.}
\end{figure}

\end{document}